\def\year{2021}\relax
\documentclass[letterpaper]{article} % DO NOT CHANGE THIS
\usepackage{aaai21}  % DO NOT CHANGE THIS
\usepackage{times}  % DO NOT CHANGE THIS
\usepackage{helvet} % DO NOT CHANGE THIS
\usepackage{courier}  % DO NOT CHANGE THIS
\usepackage[hyphens]{url}  % DO NOT CHANGE THIS
\usepackage{graphicx} % DO NOT CHANGE THIS
\usepackage{natbib}  % DO NOT CHANGE THIS AND DO NOT ADD ANY OPTIONS TO IT
\usepackage{caption} % DO NOT CHANGE THIS AND DO NOT ADD ANY OPTIONS TO IT
\usepackage{float}
\usepackage{multirow}
\usepackage{amsmath,amsthm}
\usepackage{amsfonts,color}
\usepackage{subfigure}
\usepackage{xcolor}
\usepackage[linesnumbered,noend,ruled]{algorithm2e}

\DeclareMathOperator{\nb}{Nb}

\renewcommand{\>}{\rangle}
\DeclareMathOperator{\nop}{noop}
\DeclareMathOperator{\mov}{mov}
\DeclareMathOperator{\oa}{oa}
\DeclareMathOperator{\pth}{p}
\DeclareMathOperator{\feas}{feasibleActions}
\DeclareMathOperator{\next}{nextActions}

\newcommand{\bs}{\boldsymbol}

\DeclareMathOperator{\psdd}{psdd}
\DeclareMathOperator{\sdd}{sdd}

\DeclareMathOperator{\nz}{NZ}
\DeclareMathOperator{\spset}{spset}

\DeclareMathOperator{\spath}{spath}

\DeclareMathOperator{\sset}{sset}

\DeclareMathOperator{\smset}{smset}

\newcommand\mydots{\makebox[1em][c]{.\hfil.\hfil.}}

\newtheorem{lemma}{Lemma}
\newtheorem{defi}{Definition}

\title{Combining Propositional Logic Based Decision Diagrams with Decision Making \\ in Urban Systems}
\author{
Jiajing Ling,\thanks{Equal Contribution}
Kushagra Chandak,\footnotemark[1]
Akshat Kumar\\
}
\affiliations
{
School of Information Systems\\
Singapore Management University\\
\{jjling.2018, kushagrac, akshatkumar\}@smu.edu.sg
}

\begin{document}

\maketitle

\begin{abstract}
Solving multiagent problems can be an uphill task due to uncertainty in the environment, partial observability, and scalability of the problem at hand. Especially in an urban setting, there are more challenges since we also need to maintain safety for all users while minimizing congestion of the agents as well as their travel times. To this end, we tackle the problem of multiagent pathfinding under uncertainty and partial observability where the agents are tasked to move from their starting points to ending points while also satisfying some constraints, e.g., low congestion, and model it as a multiagent reinforcement learning problem.  We compile the domain constraints using propositional logic and integrate them with the RL algorithms to enable fast simulation for RL.
\end{abstract}

\section{Introduction}
The emergence and continued rise of autonomous and semi-autonomous vehicles in the urban landscape has made its way to a number of areas for transportation and mobility like self-driving cars and delivery trucks, railways, unmanned aerial vehicles, delivery drones fleet etc. Several key challenges remain to manage such agents like maintaining safety (no collisions among vehicles), avoiding congestion and minimizing travel time to better serve the users and reduce pollution. To model such scnarios, we leverage cooperative sequential multiagent decision making, where agents acting in a partially observable and uncertain environment are required to take coordinated decisions towards a long term goal~\cite{Weiss13b}. Decentralized partially observable MDPs (Dec-POMDPs) provide a rich framework for multiagent planning~\cite{bern02,Oliehoek2016}, and are applicable in domains such as vehicle fleet optimization~\cite{NguyenKL17}, cooperative robotics~\cite{Amato2019}, and multiplayer video games~\cite{qmix18}. However, scalability remains a key challenge with even a 2-agent Dec-POMDP NEXP-Hard to solve optimally~\cite{bern02}. To address the challenge of scalability, several frameworks have been introduced that model restricted class of interactions among agents such as transition independence~\cite{Becker2004,Nair2005}, event driven and population-based interactions~\cite{Becker2004a,Varakantham2012}. Recently, several multiagent reinforcement learning (MARL) approaches are developed that push the scalability envelop~\cite{Lowe2017,Foerster2018,qmix18} by using simulation-driven optimization of agent policies.

\begin{figure}[t]
	\centering
	\includegraphics[scale=0.5]{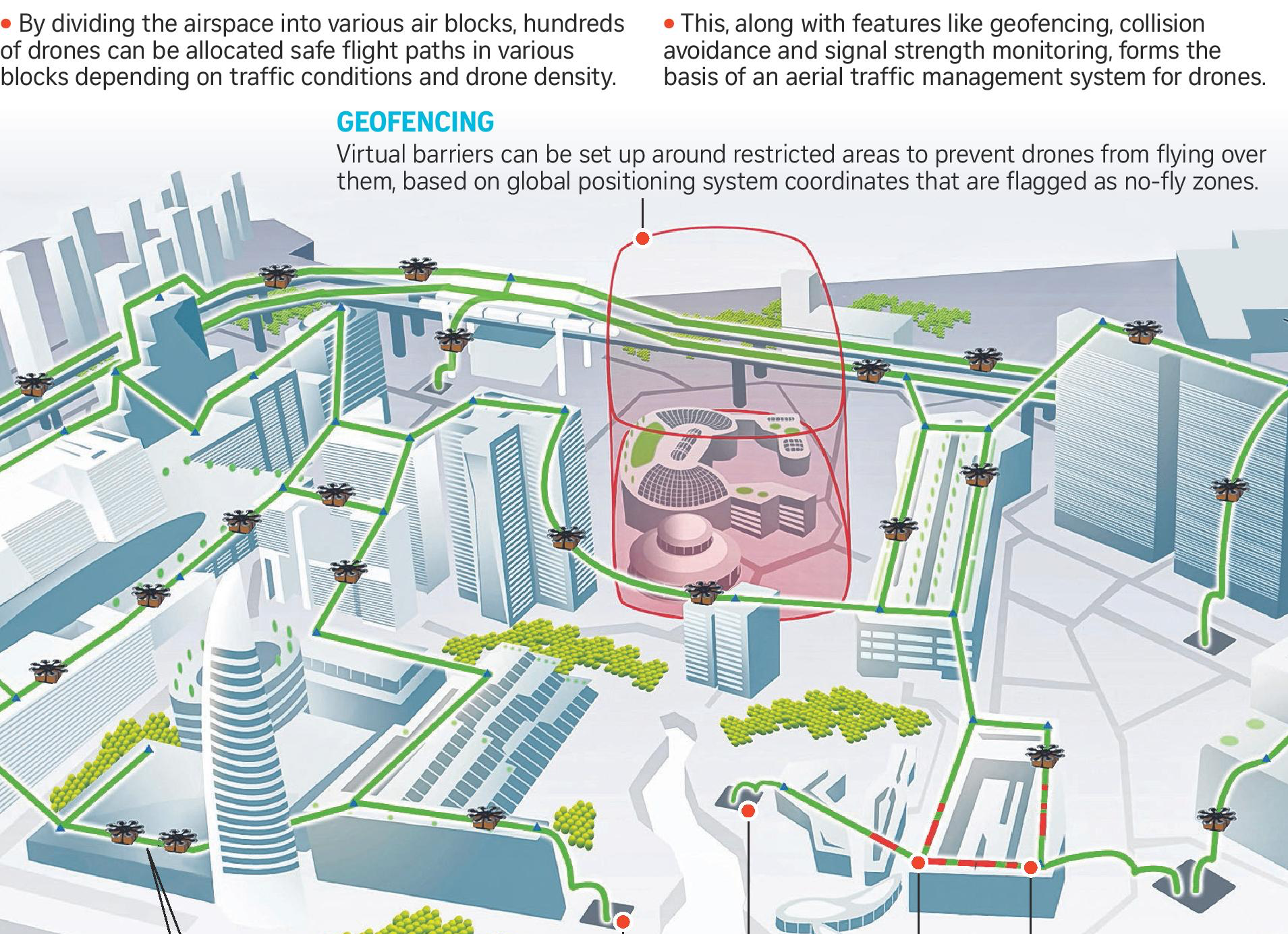}	
%	\vskip -10pt
	\caption{\small Airspace management for drone traffic~\cite{airspace}}
	\label{fig:drone}
\end{figure}

Key limitations of several MARL approaches include sample inefficiency, and difficulty in learning when rewards are sparse, which is often the case in problems with combinatorial flavor. We address such a combinatorial problem of multiagent path finding (MAPF) under uncertainty and partial observability. Even the deterministic MAPF setting where multiple agents need to find collision-free paths from their respective sources to destinations in a shared environment is NP-Hard~\cite{Yu2013}.

The MAPF problem is a general formulation that is capable of addressing several applications in the domain of urban mobility like autonomous vehicle fleet optimization~\cite{Ling0K20, primal}, taxiway path planning for aircrafts~\cite{li2019scheduling}, and train rescheduling~\cite{flat}. Figure~\ref{fig:drone} shows the airspace of a city divided into multiple geofenced airblocks. Such structured airspace can be used by drones to safely travel to their destinations~\cite{Ling0K20}. Since such spaces can have a lot of constraints, they can be modelled using our framework to manage the traffic. Deep RL approaches have been applied to MAPF under uncertainty and partial observability~\cite{primal,Ling0K20}. A key challenge faced by RL algorithms is that it takes several simulations to find even a single route to destination as model-free RL does not explicitly exploits the underlying graph connectivity. Furthermore, agents can move in cycles, specially during initial training episodes, which makes the standard RL approaches highly sample inefficient. Recent approaches combine underlying graph structure with deep neural nets for combinatorial problems such as minimum vertex cover and traveling salesman problem~\cite{Dai2017,Bello2019}. However, the knowledge compilation framework that we present provides much more explicit domain knowledge to RL approaches for MAPF. 

To address the challenges of delayed rewards, and difficulty of finding feasible routes to destinations, we compile the graph over which agents move in MAPF using propositional logic based \textit{probabilistic sentential decision diagrams} ($\psdd$)~\cite{Kisa2014}. A $\psdd$ represents probability distributions defined over the models of a given propositional theory. We use $\psdd$ to represent distribution over all simple paths (without loops) for a given source-destination pair. A key benefit is that any random sample from a $\psdd$ is gauranteed to be a valid simple path from the given source to destination. Furthermore, $\psdd$ are also equipped with associated inference methods~\cite{Shen2016} (such as computing conditional probabilities) that significantly aid RL methods (e.g., given the current partial path, what are the possible next edges that are guaranteed to lead to the destination via a simple path). Using $\psdd$ significantly helps in pruning the search space, and generate high quality training samples for the underlying learning algorithm. However, integrating $\psdd$ with different RL methods is challenging, as the standard $\psdd$ inference methods are too slow to be used in the simulation-driven RL setting where one needs to query $\psdd$ at each time step. Therefore, we also develop highly efficient $\psdd$ inference methods that specifically aid RL by enabling fast sampling of training episodes, and are more than an order of magnitude faster than generic $\psdd$ inference. Given that number of paths between a source-destination can be exponential, we also use hierarchical decomposition of the graph to enable a tractable $\psdd$ representation~\cite{Choi2017}.

To summarize, our main contributions are as follows. \textit{First}, we compile static domain information such as underlying graph connectivity using $\psdd$ for the MAPF problem under uncertainty and partial observability. \textit{Second}, we develop techniques to integrate such decision diagrams within diverse deep RL algorithms based on policy gradient and Q-learning. \textit{Third}, we develop fast algorithms to query compiled decision diagrams to enable fast simulation for MARL. We integrate our $\psdd$-based framework with previous MARL approaches~\cite{primal,Ling0K20}, and show that the resulting algorithms significantly outperform the original algorithms both in terms of sample complexity and solution quality on a number of instances. We also highlight that $\psdd$ is a general framework for incorporating constraints in decision making, and discuss extensions of the standard MAPF that can be addressed using $\psdd$.

\section{The Dec-POMDP Model and MAPF}

A Dec-POMDP is defined using the tuple $\<S, A, T, O, Z, r, n, \gamma\>$. There are $n$ agents in environment (indexed using $i\!=\!1\!:\!n$). The environment can be in one of the states $s\in S$. At each time step, agent $i$ chooses an action $a^i\in A$, resulting in the joint action $\bs{a}\in \bs{A}\equiv A^n$. As a result of the joint action, the environment transitions to a new state $s'$ with probability $T(s, \bs{a}, s')$. The joint-reward to the agent team is given as $r(s, \bs{a})$. The reward discount factor is $\gamma < 1$.

We assume a partially observable setting in which agent $i$'s observation $z^i\in Z$ is generated using the observation function $O(\bs{a}, s', z^i) = P(z^i | \bs{a}, s')$ where the last joint action taken was $\bs{a}$, and the resulting state was $s'$ (for simplicity, we have assumed the observation function is the same for all agents). As a result, different agents can receive different observations from the environment.

An agent's policy is a mapping from its action-observation history $\tau^i\in (Z\times A)^*$ to actions or $\pi^i(a^i | \tau^i; \theta^i)$, where $\theta^i$ parameterizes the policy. Let the discounted future return be denoted by $R_t \!=\! \sum_{k=0}^\infty \gamma^k r_{k+t}$. The joint-value function induced by the joint-policy of all the agents is denoted as $V^{\bs{\pi}}(s_t) \!=\! \mathbb{E}_{s_{t+1:\infty}, \bs{a}_{t:\infty}}\big[ R_t | s_t, \bs{a}_t\big]$, and joint action-value function as $Q^{\bs{\pi}}(s_t, \bs{a}_t)=\mathbb{E}_{s_{t+1:\infty}, \bs{a}_{t+1:\infty}}\big[ R_t | s_t, \bs{a}_t\big]$. The goal is to find the best joint-policy $\bs{\pi}$ to maximize the value for the starting belief $b_0$: $V(\bs{\pi})=\sum_{s} b_0(s) V^{\bs{\pi}}(s)$.

\vskip 2pt
\noindent{\textbf{Learning from simulation: }}In the RL setting, we do not have access to transition and observation functions $T$, $O$. Instead, multiagent RL approaches (MARL) learn via interacting with the environment simulator. The simulator, given the joint-action input $\bs{a}_t$ at time $t$, provides the next environment state $s_{t+1}$, generates observation $z_{t+1}^i$ for each agent, and provides the reward signal $r_t$. Similar to several previous MARL approaches, we assume a centralized learning and decentralized policy execution~\cite{Foerster2018,Lowe2017}. During centralized training, we assume access to extra information (such as environment state, actions of different agents) that help in learning value functions $V^{\bs{\pi}}$, $Q^{\bs{\pi}}$. However, during policy execution, agents rely on their local action-observation history. An agent's policy $\pi^i$ is typically implemented using recurrent neural nets to condition on action-observation history~\cite{Hausknecht2015}. However, our developed results are not affected by a particular implementation of agent policies.

\vskip 2pt
\noindent{\textbf{MARL for MAPF: }}MAPF can be mapped to a Dec-POMDP instance in multiple ways to address different variants~\cite{MaDelay17,primal,Ling0K20}. We therefore present the MAPF problem under uncertainty and partial observability using minimal assumptions to ensure the generality of our knowledge compilation framework. There is a graph $G=(V, E)$ where the set $V$ denotes the locations where agents can move, and edges connect different locations. An agent $i$ has a start vertex $s_i$ and final goal vertex $g_i$. At any time step, an agent can be located at a vertex $v\in V$, or in-transit on an edge $(u, v)$ (i.e., moving from vertex $u$ to $v$).

An agent's action set is denoted by $A=A_{\mov} \cup A_{\oa}$. Intuitively, $A_{\mov}$ denotes actions that intend to change the location of agent from the current vertex to a neighboring directly connected vertex in the graph (e.g., move up, right, down, left in a grid graph). The set $A_{\oa}$ denotes other actions that do not intend to change the location of the agent (e.g., $\nop$ that intends to make agent stay at the current vertex). Note that we do not make any assumptions regarding the actual transition after taking the action (i.e., move/stay actions may succeed or fail as per the specific MAPF instance).

Depending on the states of all the agents, an agent $i$ receives observation $z^i$. We assume that an agent is able to fully observe its current location (i.e., the vertex it is currently located at). Other information can also be part of the observation (e.g., location of agents in the local neighborhood of the agent), but we make no assumptions about such information. We make no specific assumptions about the joint-reward $r$, other than assuming that an agent prefers to reach its destination as fast as possible if the agent's movement do not conflict with other agents' movements. Typical examples of reward $r$ include penalty for every time step an agent is not at its goal vertex, positive reward at the goal vertex, a high penalty for creating congestion at vertices or edges of the graph~\cite{Ling0K20}, or  for blocking other agents from moving to their destination~\cite{primal}.

%We also assume that each agent knows about the graph structure. 

\section{Incorporating Compiled Knowledge in RL}

\begin{figure}[t]
	\centering
	\subfigure[]{\includegraphics[scale=0.25]{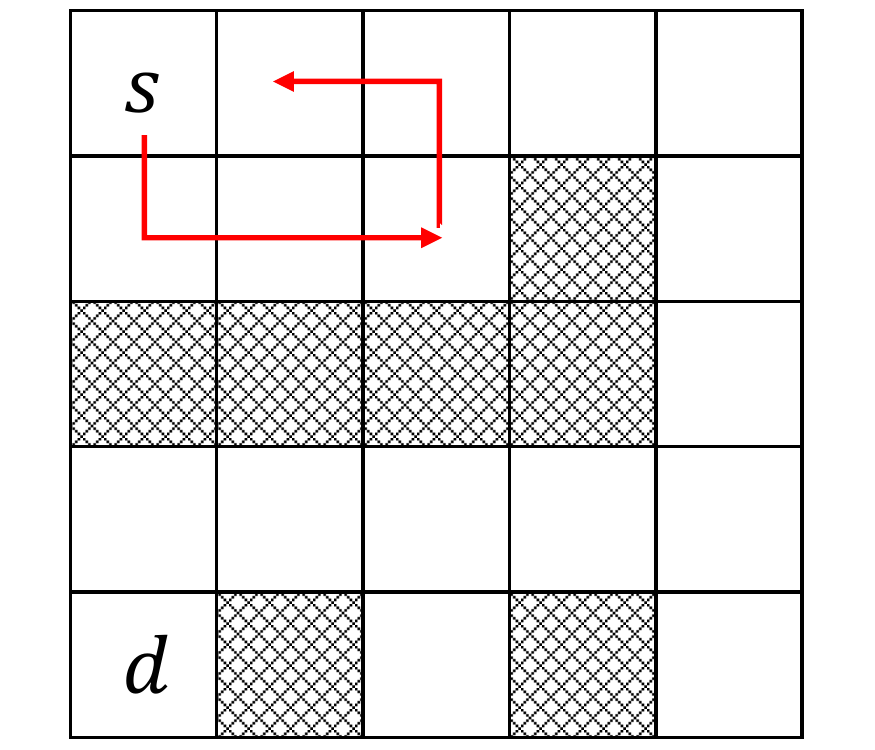}}
	\subfigure[]{\includegraphics[scale=0.25]{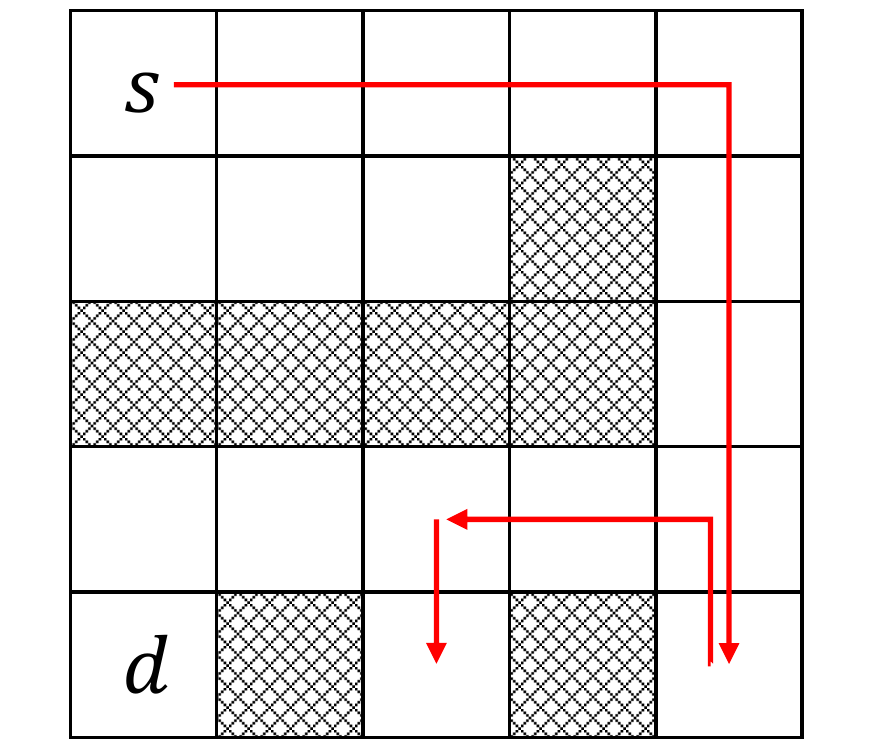}}
	\vskip -10pt
	\caption{\small Undesirable path samples for MAPF. (a) Path with a loop; (b) Path to a deadend. Dark nodes are blocked.}
	\label{fig:paths}
\end{figure}

A key challenge for RL algorithms for MAPF is that often finding feasible paths to destinations require a large number of samples. For example, figure~\ref{fig:paths}(a) shows the case when an agent loops back to one of its earlier vertex. Figure~\ref{fig:paths}(b) shows another scenario where an agent moves towards a deadend.  Such scenarios increase the training episode length in RL. Our key intuition is to develop techniques that ensure that RL approaches only sample paths that are (i) simple, (ii) always originate at the source vertex $s_i$ and end at the goal vertex $g_i$ for any agent $i$.

Let $\pth_t^i$ denote the path taken by an agent $i$ until time $t$ (or the sequence of vertices visited by an agent starting from source $s_i$). We also assume that it does not contain any cycle.
This information can be extracted from agent's history $\tau^i_t$. Let $a \!\in\! A_{\mov}$ be a movement action towards vertex $a_v$.
We assume the existence of a function $\feas(\pth_t; s_i,d_i)$ that takes as input an agent's current path $\pth_t$ and returns the set $\next = \{a \!\in\! A_{\mov} \text{ s.t. } [\pth_t, a_v]\!\rightsquigarrow\! d_i\}$. The condition $[\pth_t, a_v]\!\rightsquigarrow\! d_i$ implies there exists at least one simple path from source $s_i$ to destination $d_i$ that includes the path segment $[\pth_t, a_v]$. Thus, starting with $\pth_0 = [s_i]$, the RL approach would only sample simple paths that are guaranteed to reach an agent's destination, thereby significantly pruning the search space, and resulting in trajectories that have good potential to generate high rewards. 
%The logic behind the function $\feas$ is based on decision diagrams and inference methods presented in the next section. 
The information required for implementing $\feas$ can be compiled offline even before training and execution of policy starts (explained in next section, using  decision diagrams), and does not include any communication overhead during policy execution.
Using this abstraction, we next present simple and easy-to-implement modifications to a variety of deep multiagent RL algorithms.

\vskip 2pt
\noindent\textbf{{Policy gradient based MARL: }}We first provide a brief background of policy gradient approaches for single agent case~\cite{Sutton2000}. An agent's policy $\pi^\theta$ is parameterized using $\theta$. The policy is optimized using gradient ascent on the total expected reward $V(\theta) = \mathbb{E}_{\pi^\theta}[R_0]$. The gradient is given as:
\begin{align}
	\nabla_\theta V(\theta) = \mathbb{E}_{s_{0:\infty}, a_{0:\infty}} \Big[ \sum_{t=0}^\infty R_t \nabla_{\theta} \log \pi^\theta(a_t | s_t) \Big]
\end{align}
Above gradient expression is also extendible to the multiagent case in an analogous manner~\cite{Peshkin2000,Foerster2018}. In multiagent setting, we can compute gradient of the joint-value function $V$ w.r.t. an agent $i$'s policy parameters $\theta^i$ or $\nabla_{\theta^i}V$. The expectation is w.r.t. the joint state-action trajectories $ \mathbb{E}_{s_{0:\infty}, \bs{a}_{0:\infty}}$, and $R_t$ denotes future return for the agent team. The input to policy are some features of the agent's observation history or $\phi(\tau^i)$. The function $\phi$ can be either hard-coded (e.g., only last two observations), or can be learned using recurrent neural networks. 

For using compiled knowledge using the function $\feas$, the only change we require is in the structure of an agent's policy $\pi$ (we omit superscript $i$ for brevity). The main challenge is addressing the variable sized output of the policy in a differentiable fashion. Assuming a deep neural net based policy $\pi$, given the discrete action space $A$, the last layer of the policy has $|A|$ outputs using the softmax layer (to normalize action probabilities $\pi(a|\cdot)$). However, when using $\feas$, the probability of actions not in $\feas$ needs to be zero. However, the set $\feas$ changes as the observation history $\tau$ of the agent is updated. Therefore, a fixed sized output layer appears to create difficulties. However, we propose an easy fix. We use $\tilde{\pi}$ to denote the standard way policy $\pi$ is constructed with last layer having fixed $|A|$ outputs. However, we do not require the last layer to be a softmax layer. Instead, we re-define the policy $\pi$ as:
\vskip -1pt
{\small{
\begin{align}
\pi(a | \tau) \!=\!\! \begin{cases} 0 \text{ if } a \notin \feas(\pth(\tau); s,d) \\
					 		        \text{else } \frac{\exp\big(\tilde{\pi}(a | \phi(\tau))\big)}{\sum_{a'\in \feas(\pth(\tau); s,d)} \exp\big(\tilde{\pi}(a' | \phi(\tau))\big) }
				      \end{cases}
\end{align}}}
where $\pth(\tau)$ denotes the path taken by the agent so far, and $s,d$ are its source and destination. Sampling from $\pi$ guarantees that invalid actions are not sampled. Furthermore, $\pi$ is differentiable even when $\feas$ gives different length outputs at different time steps. The above operation can be easily implemented in autodiff libraries such as Tensorflow without requiring a major change in the policy structure $\pi$. 

\vskip 2pt
\noindent\textbf{{Q-learning based MARL: }}Deep Q-learning for the single agent case~\cite{Volodymyr2015} has been extended to the multiagent setting also ~\cite{qmix18}. In the QMIX approach~\cite{qmix18}, the joint action-value function $Q_{tot}(\bs{\tau}, \bs{a};\psi)$ is factorized as (non-linear) combination of action-value functions $Q_i(\tau^i, a^i; \theta^i)$ of each agent $i$. A key operation when training different parameters $\theta^i$ and $\psi$ involves maximizing $\max_{\bs{a}}Q_{tot}(\bs{\tau}, \bs{a};\phi)$ (for details we refer to~\citeauthor{qmix18}). This operation is intractable in general, however, under certain conditions, it can be approximated by maximizing individual Q functions $\max_{a\in A} Q_i(\tau^i, a^i)$ in QMIX. We require two simple changes to incorporate our knowledge compilation scheme in QMIX. First, instead of maximizing over all the actions, we maximize only over feasible actions of an agent as $\max_{a\in \feas{\pth(\tau^i; s^i, d^i)}} Q_i(\tau^i, a)$. Second, in Q-learning, typically a replay buffer is also used which stores samples from the environment as $(\bs{\tau}, \bs{a}, \bs{\tau}^\prime, r)$. In our case, we also store additionally the set of  feasible actions for the next observation history $\tau^{\prime i}$ for each agent $i$ as $\feas(\pth(\tau^{\prime i}); s^i, d^i)$ along with the tuple $(\bs{\tau}, \bs{a}, \bs{\tau}^\prime, r)$. The reason is when this tuple is \textit{replayed}, we have to maximize $Q^i(\tau^{\prime i}, a)$ over $a \in \feas({\pth(\tau^{\prime i}); s^i, d^i)}$, and storing the set $\feas({\pth(\tau^{\prime i}); s^i, d^i)}$ would reduce computation.

We have integrated our knowledge compilation framework with two policy gradient approaches proposed in~\cite{primal,Ling0K20} (one using feedforward neural net, another using recurrent neural network based policy), and a QMIX-variant~\cite{Fu2019} for MAPF, demonstrating the generalization power of the framework for a range of MARL solution methods.

\begin{figure*}[t]
	\centering
	\subfigure[]{\includegraphics[width=4cm, height=3cm]{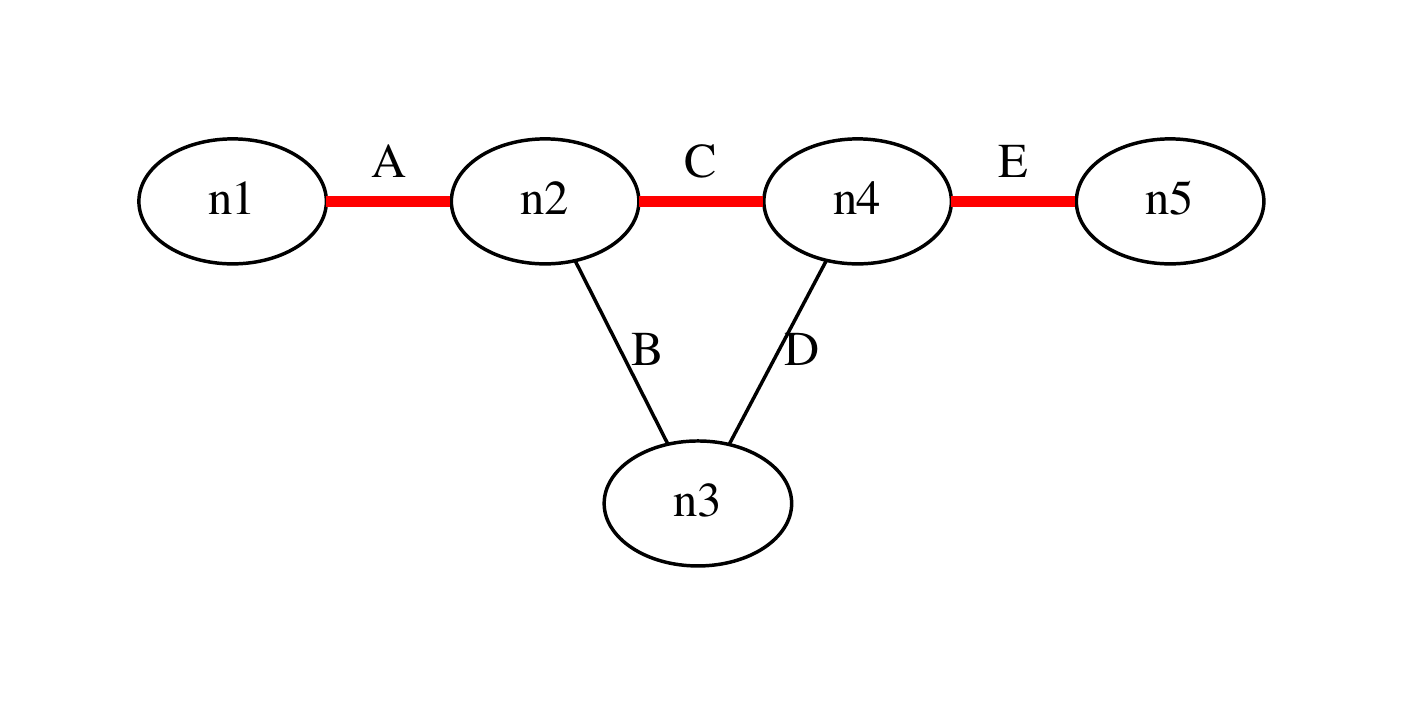}}
	\subfigure[]{\includegraphics[width=4cm, height=5.5cm]{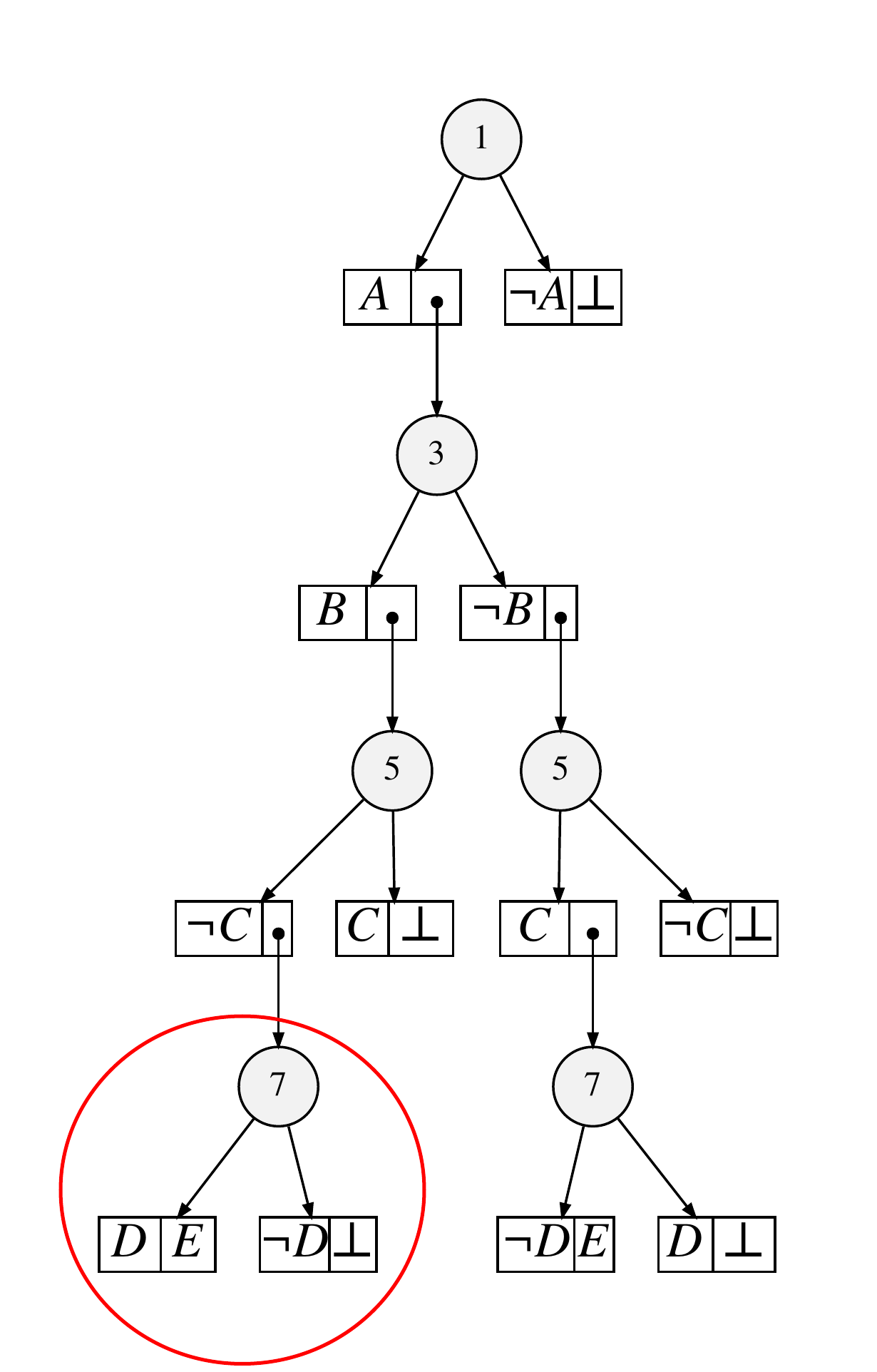}}
	\subfigure[]{\includegraphics[width=3.5cm, height=5.5cm]{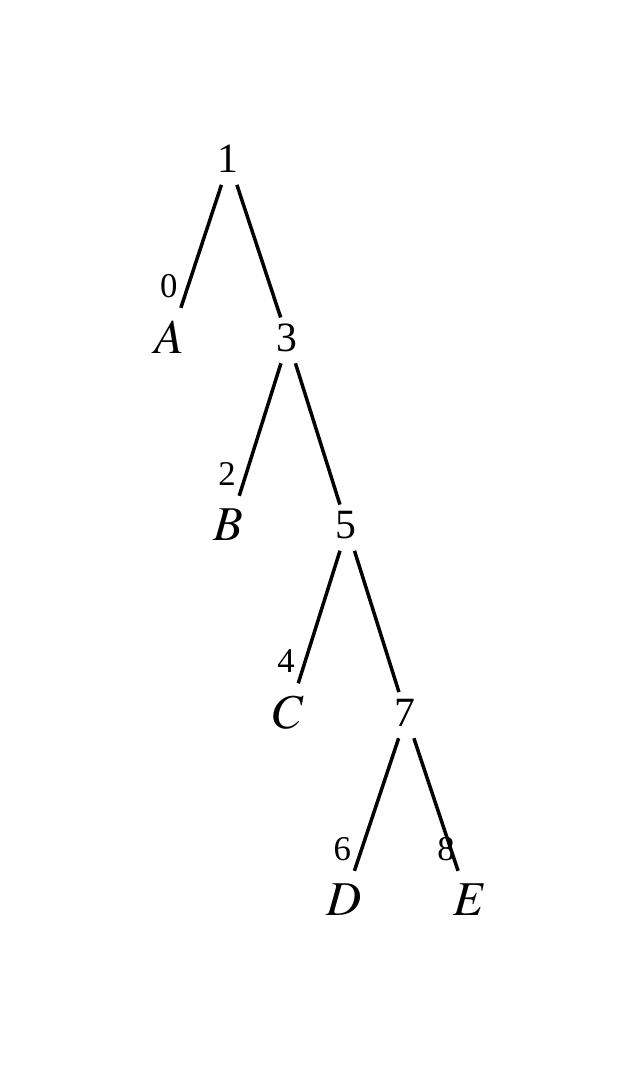}}
	\subfigure[]{\includegraphics[width=4cm, height=5.5cm]{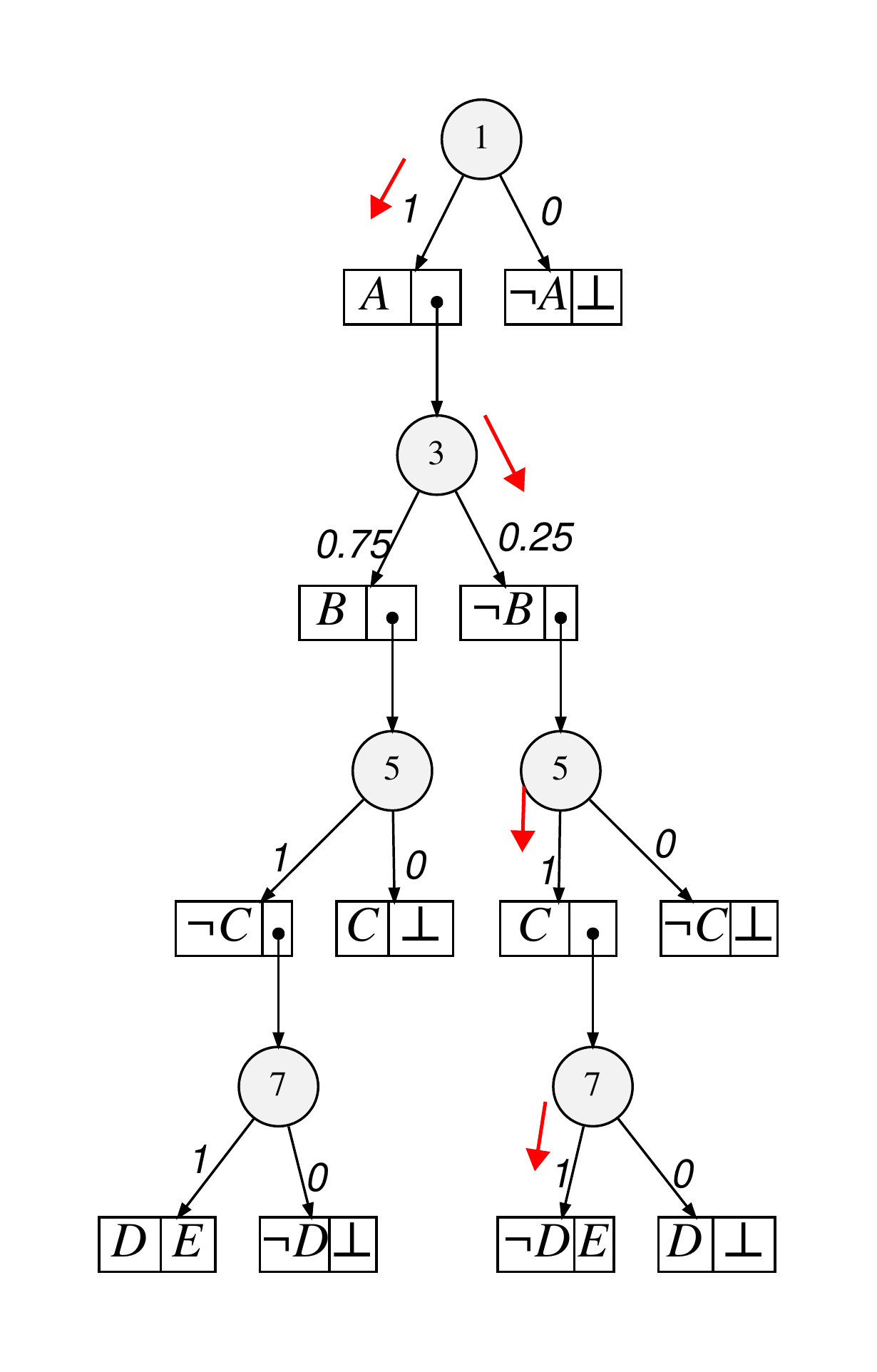}}

	\vskip -10pt
	\caption{\small (a) A simple path in a graph from $s=n1$ to $d=n5$ is highlighted in red and can be written as a propositional sentence $A\land C\land E\land\neg B\land\neg D$; (b) An $\sdd$ for the graph in (a) where the encircled node represents a decision node $(p_1,s_1),(p_2,s_2)$ (c) a \textit{right-linear} vtree for the $\sdd$; (d) $\psdd$ with parameters annotated on decision nodes.} 
% 	To sample a path, we traverse the $\psdd$ top-down selecting one of the branches in each decision node and selecting both the prime and the sub for that branch.}
	\label{fig:sdd}
\end{figure*}

% \vskip 2pt
\section{Compiling and Querying Decision Diagrams for MAPF}
% \textbf{Notation: }A Boolean random variable is denoted by uppercase letters, e.g., $X$. A lowercase letter $x$ denotes an assignment of true or false to $X$. $X$ or $\neg X$ are called literals and denote the variable and its negation, respectively. True and False are represented by symbols $\top$ and $\bot$. Sets of variables $\mathbf{X}$ and joint assignments $\mathbf{x}$ are denoted in bold. An assignment $\mathbf{x}$ that satisfies logical sentence $\alpha$ is denoted by $\mathbf{x} \models \alpha$.
% Before we introduce $\sdd$s, we formalize the notion of {\it structured} spaces [Tractable Learning for Structured Probability Spaces].
% \paragraph{Def.}Consider a set of $n$ Boolean variables $\mathbf{X} = \{X_1,\mydots,X_n\}$. The space of all $2^n$ instantiations of $X$ is called an {\it unstructured} space. If the space respects a given set of logical constraints which doesn't allow some instantiations, it's called a {\it structured} space.
We now present our decision diagram based approach to implement the $\feas$ function. 
Let upper case letters ($X$) denote variables and lowercase letters ($x$) denote their instantiations. Bold upper case letter ($\textbf{X}$) denotes a set of variables and their lower case counterparts ($\textbf{x}$) denote the instantiations.

\noindent{\textbf{Paths as a Boolean formula: }}A path $\pth$ from a given source $s$ to the destination $d$ in the underlying undirected graph $G=(V, E)$ can be represented as a Boolean formula as follows. Consider Boolean random variables $X_{i,j}$ for each edge $(i,j)\in E$. If an edge $(i,j)$ occurs in $\pth$, then $X_{i,j}$ is set to true, otherwise it's set to false. Hence, conjunction of these {\it literals} denotes path $\pth$, and the Boolean formula representing \textit{all paths} is obtained by simply disjoining formulas for all such paths~\cite{Choi2016StructuredFI}. An example path in a graph is given in fig~\ref{fig:sdd}(a).

%To compile the graph connectivity in MAPF using propositional logic based decision diagrams, we first show how a path in the underlying  graph $G=(V, E)$ can be represented as a Boolean formula. Consider Boolean random variables $X_{i,j}$ for each edge $(i,j)\in E$. If an edge $(i,j)$ occurs in a path from source node $s$ to destination node $d$ in $G$, then $X_{i,j}$ is set to true, otherwise it's set to false. Hence, conjunction of these {\it literals} is a path in $G$ and the Boolean formula representing all paths is obtained by simply disjoining all such paths~\cite{Choi2016StructuredFI}. An example path in a graph is given in fig~\ref{fig:sdd}(a).

% Consider an example of a 2x2 grid (fig.~\ref{fig:sdd}(a)) in which we represent a Boolean formula for all simple (no cycle) paths from $s=1$ to $d=4$ by $(X_{1,2}\land X_{2,4}\land\neg X_{1,3}\land\neg X_{3,4}) \lor (\neg X_{1,2}\land \neg X_{2,4}\land X_{1,3}\land X_{3,4})$.

\noindent{\textbf{Sentential decision diagrams: }}Since the number of paths between two nodes can be exponential, we need a compact representation of the Boolean formula representing paths. To this end, we use \textit{sentential decision diagram} or $\sdd$~\cite{darwiche2011sdd}. It is a Boolean function $f(\textbf{X},\textbf{Y})$ on some non-overlapping variable sets $\textbf{X},\textbf{Y}$ and is written as a \textit{decomposition} in terms of functions on $\textbf{X}$ and $\textbf{Y}$. In particular, $f=(p_1(\textbf{X})\land s_1(\textbf{Y}))\lor\mydots\lor(p_n(\textbf{X})\land s_n(\textbf{Y}))$, with each \textit{element} $(p_i,s_i)$ of the decomposition composed of a \textit{prime} $p_i$ and a \textit{sub} $s_i$. A $\sdd$ represented as a decision diagram describes members of a combinatorial space (e.g., paths in a graph) using propositional logic in a tractable manner. It has two kinds of nodes: 
\begin{itemize}
	\item[-] \textit{terminal node}, which can be a literal ($X$ or $\neg X$), always true ($\top$) or always false ($\bot$), and
	\item[-] \textit{decision node}, which is represented as $(p_1\land s_1)\lor\mydots\lor(p_n\land s_n)$ where all $(p_i,s_i)$ pairs are recursively $\sdd$s and the primes are always consistent, mutually exclusive and exhaustive.
\end{itemize}

Figure~\ref{fig:sdd}(b) represents an $\sdd$ for the graph in fig~\ref{fig:sdd}(a) encoding all paths from n1 to n5. The encircled node is a decision node with two elements $(D,E)$ and $(\neg D,\bot)$. The primes are $D$ and $\neg D$ and the subs are $E$ and $\bot$. The Boolean formula representing this $\sdd$ node is $(D\land E)\lor(\neg D\land\bot)$ which is equivalent to $D\land E$. The Boolean formula encoded by the whole $\sdd$ is given by the root node of the $\sdd$.% which is normalized for the root node of the corresponding vtree (in figure~\ref{fig:sdd}(c)).

An $\sdd$ is characterized by a \textit{full} binary tree, called a \textit{vtree}, which induces a total order on the variables from a left-right traversal of the vtree. E.g., for the vtree in figure~\ref{fig:sdd}(c), the variable order is $(A, B, C, D, E)$. Given a  fixed vtree, the $\sdd$ is unique. 
An $\sdd$ node $n$ is \textit{normalized} (or associated with) for a vtree node $v$ as follows: 
\begin{itemize}
	\item[-] If $n$ is a terminal node, then $v$ is a leaf vtree node which contains the variable of $n$ (if any).
	\item[-] If $n$ is a decision node, then $n$'s primes (subs) are normalized for the left (right) child of $v$.
	\item[-] If $n$ is the root node, then $v$ is the root vtree node.
	%$v$ is an internal vtree node with its left child $v^l$ containing primes of $n$ and its right child $v^r$ containing subs of $n$.
\end{itemize}
{Intuitively, a decision node $n$ being normalized for vtree node $v$ implies that the Boolean formula encoded by $n$ contains only those variables contained in the sub-tree rooted at $v$.} We will use this normalization property for our analysis later. 
The Boolean formula encoding the domain knowledge can be compiled into a decision diagram using the $\sdd$ compiler~\cite{oztok2015top}. The resulting $\sdd$ may not be exponential in size even though it is representing an exponential number of objects. 

\vskip 2pt
\noindent{\textbf{Probabilistic Sentential decision diagrams: }}In our case, for computing $\feas$, we also need to associate a probability distribution with the $\sdd$ that encodes all the paths from a given source to destination. The key benefit is that we can exploit associated inference methods such as computing conditional probabilities, which will help in computing $\feas$.

If we parameterize each of the decision nodes of the $\sdd$, such that the local parameters form a distribution, the resulting probabilistic structure is called a $\psdd$ or a \textit{probabilistic}~$\sdd$~\cite{Kisa2014}. It can be used to represent discrete probability distributions $Pr(\textbf{X})$ where several instantiations $\textbf{x}$ have zero probability $Pr(\textbf{x})=0$ because of the constraints imposed on the space. More concretely, a $\psdd$ \textit{normalized} for an $\sdd$ is defined as follows:
\begin{itemize}
	\item[-] For each decision node $(p_i,s_i),\mydots,(p_n,s_n)$, there's a positive parameter $\theta_i$ such that $\sum_{i=1}^{n}\theta_i=1$ and $\theta_i=0$ iff $s_i=\bot$.
	\item[-] For each terminal node $\top$, there's a parameter $0<\theta<1$.
\end{itemize}

\noindent{$\psdd$s} are tractable models of probability distributions as several probabilistic queries can be performed in poly-time such as computing marginal probabilities, or conditional probabilities. 

\vskip 2pt
\noindent{\textbf{$\bs{\nz}$ (Non-Zero) Inference for $\bs{\feas}$: }}Given an $\sdd$ encoding all simple paths from a source $s$ to a destination $d$, we uniformly parameterize this $\sdd$ as noted earlier. That is, for a decision node $(p_i,s_i),\mydots,(p_n,s_n)$, each $\theta_i$ is the same (except when $s_i=\bot$, then $\theta_i=0$). And we also enforce that non-zero $\theta_i$s normalize to $1$. This strategy makes sure that the probability of each simple path from $s$ to $d$ is non-zero. Assume that the current sampled path by the agent is $\pth$ (in the context of $\psdd$, we assume that $\pth$ is a set of edges in graph $G$ traversed from source $s$ by the agent). Let $v_{\pth}$ denote the current vertex of the agent (and assume $v_{\pth}$ is not the destination). Let $\nb(v_{\pth})$ denote all direct neighbors of $v_{\pth}$.  The $\feas$ set is given as:
{\small
	\begin{align}
	\feas(\pth) = \{ v'\in \nb(v_{\pth}) \wedge (v_{\pth}, v') \notin \pth \nonumber \\
	\wedge  Pr((v_{\pth}, v') | \pth) > 0  \}
	\end{align}}
That is, if the conditional probability  $Pr((v_{\pth}, v') | \pth) = 0$, then $v'$ can be pruned from the action set as it implies there is no simple path to destination $d$ that takes the edge $(v_{\pth}, v')$ after taking the path $\pth$. This strategy seems straightforward to implement as $\psdd$ is equipped with inference methods to compute conditional probabilities. However, in RL, this inference needs to be done at each time step for each training episode. We observed empirically that this method was extremely slow, and it was impractical to scale it for multiple agents. We therefore next develop our customized inference technique that is much faster than this generic inference.

\vskip 2pt
\noindent{\textbf{Sub-context connectivity analysis for $\bs{\nz}$ Inference: }}We note that all the discussion below is for a $\psdd$ that encodes all simple paths from a source $s$ to destination $d$, and the $\psdd$ is normalized for some \textit{right linear vtree}. Proofs for different results are provided in the supplementary material in the full paper available on Arxiv.

%They are efficacious in modeling distributions over combinatorial objects like rankings, permutations, trees and paths in a graph where the underlying $\sdd$ captures the combinatorial object while the parameterization of the $\psdd$ induces a tractable distribution on those objects. Domain knowledge or prior symbolic knowledge can be encoded by representing objects of a combinatorial space using $\sdd$s. $\psdd$s can also be used to represent Bayesian Networks~\cite{Shen2018ConditionalPM,shen2019structured}.

\begin{lemma}
	\label{lem:1}
In a $\psdd$ normalized for a right linear vtree, each prime is a literal ($X$ or $\neg X$) or $\top$.
\end{lemma}
The above result is a direct consequence of the manner in which the underlying $\sdd$ is constructed using a right linear vtree.

We sample a path from such a $\psdd$ by traversing it in a top-down fashion and selecting one branch at a time for each of the decision nodes according to the probability for that branch and then selecting the prime and recursively going down the sub~\cite{Kisa2014}. As all the prime nodes are terminal as per lemma~\ref{lem:1}, if the prime node is a positive literal $X$, then we select the edge $e$ corresponding to $X$ for our path (say $e_X$). If prime node is $\neg X$, then we do not select edge $e_X$. We show in the supplement that the prime nodes encountered during such sampling procedure for a $\psdd$ that encodes simple paths cannot be $\top$.

As an example, consider the graph in fig~\ref{fig:sdd}(a) and its corresponding $\psdd$ in fig~\ref{fig:sdd}(d). We start at the root of the $\psdd$ and select the left branch with probability 1. We then select the prime $A$ in our sample and recursively go down its sub as shown by the red arrows. The final sampled path is $A - C - E$ and the corresponding Boolean formula is $A\land\neg B\land C\land\neg D\land E$.

\begin{defi}(S-Path)
%Consider $\psdd$ nodes normalized for the the two deepest vtree nodes $v^d_l$ and $v^d_l$.
Let $n$ be a $\psdd$ node normalized either for $\tilde{v}_l$ and $\tilde{v}_r$, the two deepest vtree nodes. 
Let $(p_1, s_1), \mydots, (p_k, s_k)$ be the elements appearing on some path from the $\psdd$ root to node $n$ (i.e., $n=p_k$ or $n=s_k$). Then $p_1\wedge \mydots \wedge p_k \wedge n$ is called an \textbf{s-path} for node $n$, and is feasible iff $s_i \neq \bot$. 	
\end{defi}
In figure~\ref{fig:sdd}(c), $\tilde{v}_l$ is $D$, and $\tilde{v}_r$ is $E$. There can be multiple s-paths for a node $n$. Let $\spset$ denote the set of all \textit{feasible} s-paths for all $\psdd$ nodes $n$ normalized either for $\tilde{v}_l$ or $\tilde{v}_r$.

\begin{lemma}
	\label{lem:1to1}
There is a one-to-one mapping between s-paths in the set $\spset$ and the set of all simple paths in $G$ from source $s$ to destination $d$.
\end{lemma}
The above lemma states that if we find a feasible s-path in the $\psdd$, then it would correspond to a valid simple path from source $s$ to destination $d$ in the graph $G$ which will also have nonzero probability as per our $\psdd$. Reading off the path in $G$ given a feasible s-path is straightforward. A feasible s-path is also a conjunction of literals (using lemma~\ref{lem:1}, and if $n$ is a sub, it will also be a literal as $n$ is normalized for deepest node in vtree). For each positive literal $X$ in s-path, we include its corresponding edge $e_X$ in the path in $G$. The set of resulting edges would form a simple path in $G$.

This result also provides a strategy for our fast $\nz$ inference. Given a path $\pth$ in graph $G$, our goal is to find whether $Pr((v_{\pth}, v') | \pth) > 0$. If we can prove that there exists an s-path $sp\in \spset$ such that its corresponding path in graph $G$ (using lemma~\ref{lem:1to1}) contains all the edges in $\pth$ and $(v_{\pth}, v')$, then $Pr((v_{\pth}, v') | \pth)$ must be nonzero. We need few additional results below to turn this insight into an efficient algorithmic procedure.

\begin{defi}(Sub-context~\cite{Kisa2014})
Let $(p_1, s_1), \mydots, (p_k, s_k)$ be the elements appearing on some path from the $\sdd$ root to node $n$ (i.e., $n=p_k$ or $n=s_k$). Then $p_1\wedge \mydots \wedge p_k$ is called a sub-context $sc$ for node $n$, and is feasible iff $s_i \neq \bot$. 	
\end{defi}
Notice that a $\psdd$ node $n$ can have multiple (feasible) sub-context as a $\psdd$ is a directed acyclic graph (DAG). Essentially, each sub-context corresponds to one possible way of reaching node $n$ from the $\psdd$ root. For a right linear vtree, a feasible sub-context is a conjunction of literals as all primes are literals (lemma~\ref{lem:1}).

Given two $\psdd$ nodes $n$ and $n'$, we say that $n'$ is \textit{deeper} than $n$ if the vtree node $v'$ for which $n'$ is normalized is deeper than vtree node $v$ for which $n$ is normalized.

\begin{defi}(Sub-context set) 
Let $X$ be a positive literal, and let $p_{i_1}, \mydots,p_{i_k}$ be $\psdd$ prime nodes such that each $p_{i} = X$. Let $ssc_1\mydots ssc_k$ be sets such that each $ssc_j$ contains all the feasible sub-contexts of $p_{i_j}$. Then the sub-context set of $X$ denoted by $\sset(X)$ is defined as $\sset(X)=\cup_{j=1}^{k}ssc_j$
\end{defi}

% Intuitively, $\sset(X)$ denotes all paths from $\psdd$ root to some node $p_i$ such that positive literal $X$ appears in the sub-context of $p_i$. 
We now show the procedure to perform sub-context connectivity analysis for $\nz$ inference. Assume that the current sampled path in graph $G$ is $\pth = \{e_1, \mydots, e_k\}$ (each $e_i$ is the edge traversed by the agent so far). Let the current vertex of the agent be $v_{\pth}$. Let $e=(v_{\pth}, v')$ be one possible edge in $G$ that the agent can traverse next. Let $X_{e_1},\mydots, X_{e_k}, X_e$ be the respective Boolean variables for the different edges. We wish to determine whether $P(X_e | X_{e_1},\mydots, X_{e_k})$\footnote{shorthand for $P(X_e=1 | X_{e_1}=1, \mydots, X_{e_k}=1)$} is greater than zero. We follow the following steps to determine this.

\begin{enumerate}
	\item Find the variable $\tilde{X}\in \{X_{e_1}, \mydots, X_{e_k}, X_e\} $ that is deepest in the vtree order.
	\item Check if there exists a sub-context $sc\in\sset(\tilde{X})$ such that $sc$ contains all the positive literals {\small$\{X_{e_1}, \mydots, X_{e_k}, X_e\}$}. Concretely, check if $\exists$ $sc\!\in\!\sset(\tilde{X})$ s.t. $sc \!\land\! X \!=\! sc$, {\small$\forall X \!\in\! \{X_{e_1}, \mydots, X_{e_k}, X_e\}$}. Denote this sub-context $sc^*$ (if exists).
	\item 
% 	If $sc^*$ exists, then it contains all the positive literals corresponding to the edges traversed so far and the edge $e$ currently under consideration. 
	Since $sc^*$ is the sub-context of the variable deepest in the vtree order among $\{X_{e_1},\mydots, X_{e_k}, X_e\}$, it can be extended to a feasible s-path $sp\in \spset$ such that $sp$ contains $sc^*$ (or $sc^\star\land sp = sp$). (Proved formally in supplementary). Therefore, we have shown the existence of a feasible s-path $sp$ that contains all literals $\{X_{e_1},\mydots, X_{e_k}, X_e\}$, and by lemma~\ref{lem:1to1}, there also exists a simple path in graph $G$ that contains the edges $\{e_1, \mydots, e_k, e\}$. Therefore, $P(X_e | X_{e_1}, \mydots, X_{e_k})$ is non-zero.
	\item If $sc^*$ does not exist, then a feasible s-path cannot be found containing all the literals $\{X_{e_1}, \mydots, X_{e_k}, X_e\}$ (proved in supplementary). Therefore, $P(X_e | X_{e_1}, \mydots, X_{e_k})$ is zero.
\end{enumerate}
Step number 2 in the method above is computationally the most challenging. We develop additional results in the supplementary material that further optimize this step, resulting in a fast and practical algorithm for $\nz$ inference.

% We illustrate this procedure using an example. Consider the graph in fig~\ref{fig:sdd}(a) with source $s=n1$ and destination $d=n5$. The corresponding $\psdd$ and vtree is given in fig~\ref{fig:sdd}(c) and fig~\ref{fig:sdd}(d). Let the partial path be $\pth$ with $\pathEdges=\{(n1,n2),(n2,n4)\}$ and $\evi=\{A,C\}$. Now we want to find if the edge $e=(n4,n5)$ can be selected, i.e., if $Pr(X_e=E|\evi)>0$. First we find $X_D$, which turns out to be the literal $E$. We also compute $\sset(E)=\{(A\land B\land\neg C\land D), (A\land\neg B\land C\land\neg D)\}$. From $\sset(E)$, we can clearly see that $sc_X=(A\land\neg B\land C\land\neg D)$ contains all the literals in the set $\evi$. Now we see that $sc_X$ can be extended to an $\spath$ $sp=(A\land\neg B\land C\land\neg D\land E)$ since $sc_X\land sp = sp$. 

\vskip 2pt
\noindent{\textbf{Hierarchical clustering for large graphs: }}For increasing the scalability of the $\psdd$ framework and $\nz$ inference for large graphs, we take motivation from~\cite{choi2017tractability,shen2019structured}. These previous results show that by suitably partitioning the graph $G$ among clusters, we can keep the size of the $\psdd$ tractable even for very large graphs. Such partitioning does result in the loss of expressiveness as the $\psdd$ for the partitioned graph may omit some simple paths, but empirically, we found that this partitioning scheme still improved efficiency of the underlying RL algorithms significantly. This partitioning method is described in the supplementary material in more detail.

\section{Extensions and Modeling Other Logical Constraints}
The framework that we presented can be used to compile a number of different kinds of constraints. For example, the agent has to first go to a pickup location and then to a delivery location~\cite{liu2019task}, or TSP-like constraints where the agent has to visit some locations before reaching the destination while avoiding collisions. An example is explained below in more detail.

\noindent{\textbf{Landmark Constraints: }}This framework can be extended to settings where an agent is required to visit some landmarks before reaching the destination. We can construct the Boolean formula representing such a constraint by taking incident edge variables for each of the landmarks and allowing at least one of them to be true. We can then multiply~\cite{shen2016tractable} the PSDD representing such a formula with the PSDD representing simple path constraint. For example, if $n_i$ is a node representing a landmark and $A, B, C, D$ are Boolean variables representing the edges incident on $n_i$, then we can represent the constraint for $n_i$ as $\beta_i = A\lor B\lor C\lor D$. For $k$ such landmarks, we can similarly represent the constraints $\beta_1,\ldots,\beta_k$. Then the Boolean formula for all the landmarks would be $\beta = \bigwedge_{i=1}^k\beta_i$ and can be compiled as a PSDD. Now, if $\alpha$ is a PSDD representing simple paths between a source and a destination, then we can multiply $\alpha$ and $\beta$ to get the final PSDD representing simple paths where an agent is required to visit some landmarks before the destination. This strategy can be scaled up by hierarchical partitioning of the graph~\cite{choi2017tractability,shen2019structured} and can be used to represent complex constraints by multiplying them. This process is also modular since the constraints are separately modeled from the underlying graph connectivity.

 Furthermore, this framework can also be used in cases where the underlying graph connectivity is dynamic; e.g., in scenarios where edges are dynamically getting blocked over time or the graph is revealed with time like the Canadian Traveller Problem~\cite{LIAO201480}. Any observation about blocked edges at a time can become the evidence, and by conditioning on this evidence, the agent can rule out routes via such blocked edges. The generalizability and flexibility of this framework make it a promising approach in combining domain knowledge with models for RL, pathfinding, and other areas.

\section{Empirical Evaluation}

%To represent $\psdd$ and $\sdd$ in our experiments, we use the \texttt{GRAPHILLION}~\cite{inoue2016graphillion} package to first construct a ZDD and then convert it to $\sdd$~\cite{nishino2016zero,nishino2017compiling}. We also use the \texttt{PySDD}~\cite{darwiche2018recent} package for constructing $\sdd$s and \texttt{PyPSDD}\footnote{https://github.com/art-ai/pypsdd} package for constructing and doing inference on $\psdd$s.

%We now evaluate our knowledge-based framework integrated with a variety of deep-RL algorithms (policy gradient, Q-learning, actor-critic). 
We present results to show how the integration of our framework with previous multiagent deep-RL approaches based on policy gradient and Q-learning~\cite{primal,Ling0K20} performs better in MAPF problems in terms of both sample efficiency and solution quality on a number different maps with different number of agents.

% In this section, we present the results of simulation speed of \textcolor{red}{context method and posterior sampling method} in open grids with varying size.
\noindent{\textbf{Simulation Speed: }}We show comparisons between our method and $\psdd$ inference method for calculating marginal probabilities. Our approach is more than an order of magnitude faster.

\begin{table}[H]
\centering
\begin{tabular}{ |c|c c c|c| } 
 \hline
 \multirow{2}{5em}{\vspace{-0.5em} \textbf{Approach }} & \multicolumn{3}{|c|}{\textbf{No Clustering}} & \multicolumn{1}{|c|}{\textbf{Clustering}} \\ \cline{2-5}
        & \multicolumn{1}{|c|}{3x3} & \multicolumn{1}{|c|}{4x4} & \multicolumn{1}{|c|}{5x5} & \multicolumn{1}{|c|}{10x10} \\		
 \hline
SCANZ & 1.84 & 3.86 & 19.82 & 407.95\\ 
 \hline
$\psdd$ inference & 26.55 & 158.41 & 979.71 & 402665.98\\ 
 \hline
\end{tabular}
	%\vskip -10pt
	\caption{\small Simulation speed comparison (in seconds)}
	
	\label{simulation_speed}
\end{table}

%First we evaluate the sampling speed of $\psdd$ inference method for computing conditional probabilities and our approach based on sub-context connectivity analysis for $\nz$ inference (SCANZ) in open grid maps of different sizes 3x3, 4x4, and 5x5. The experiments were performed on a single desktop machine with an Intel i7-8700 CPU and 32GB RAM. For each map, the source and destination are the top right node and bottom left node respectively. We randomly generate 10,000 paths given the source and destination pairs using both SCANZ and $\psdd$ conditional probabilities and calculate the running time for the entire path simulation. Table~\ref{simulation_speed} shows that SCANZ is more than an order of magnitude faster than $\psdd$ inference.

%\begin{table}[h]
%\centering
%\begin{tabular}{ |c|c|c|c|c| } 
% \hline
% 		 & 3x3 & 4x4 & 5x5\\ 
% \hline
%SCANZ & 1.84 & 3.86 & 19.82\\ 
% \hline
%$\psdd$ inference & 26.55 & 158.41 & 979.71\\ 
% \hline
%\end{tabular}
%	\caption{\small Simulation speed comparison (in seconds)}
%	\label{simulation_speed}
%\end{table}

\begin{figure*}[t]
	\centering
	\subfigure[4x4 grid]{\includegraphics[scale=0.285]{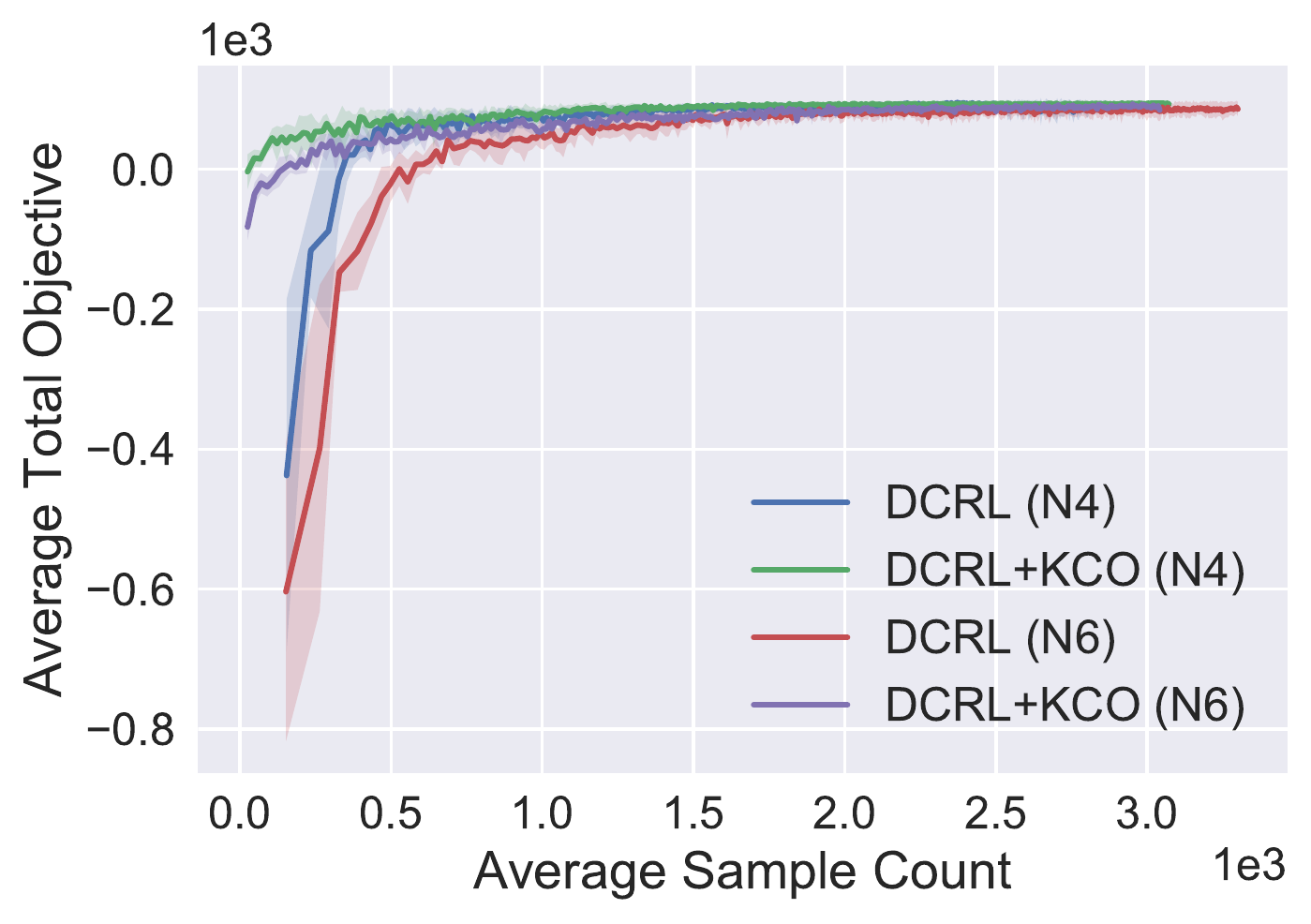}}
	\subfigure[8x8 grid]{\includegraphics[scale=0.285]{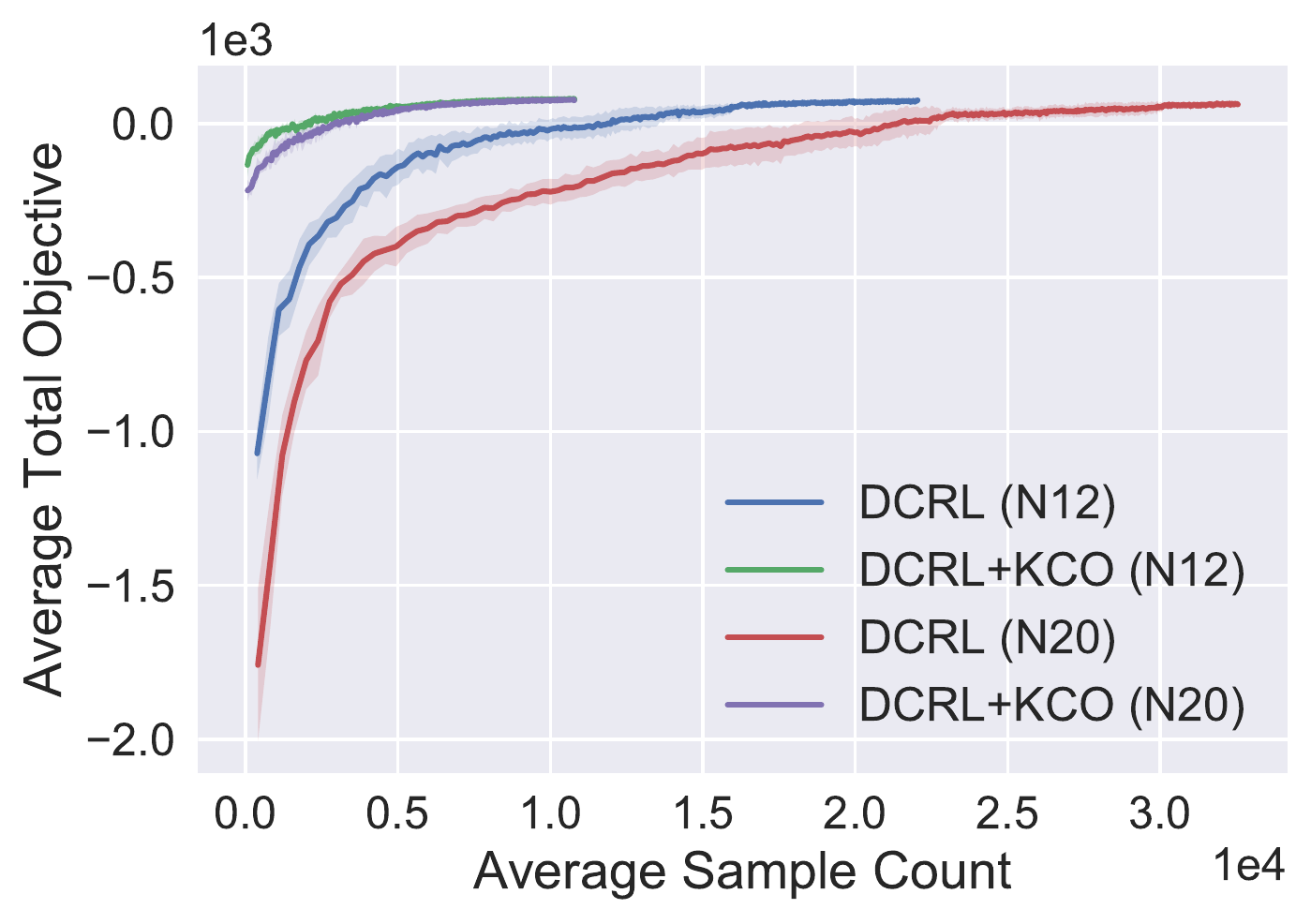}}
	\subfigure[10x10 grid]{\includegraphics[scale=0.285]{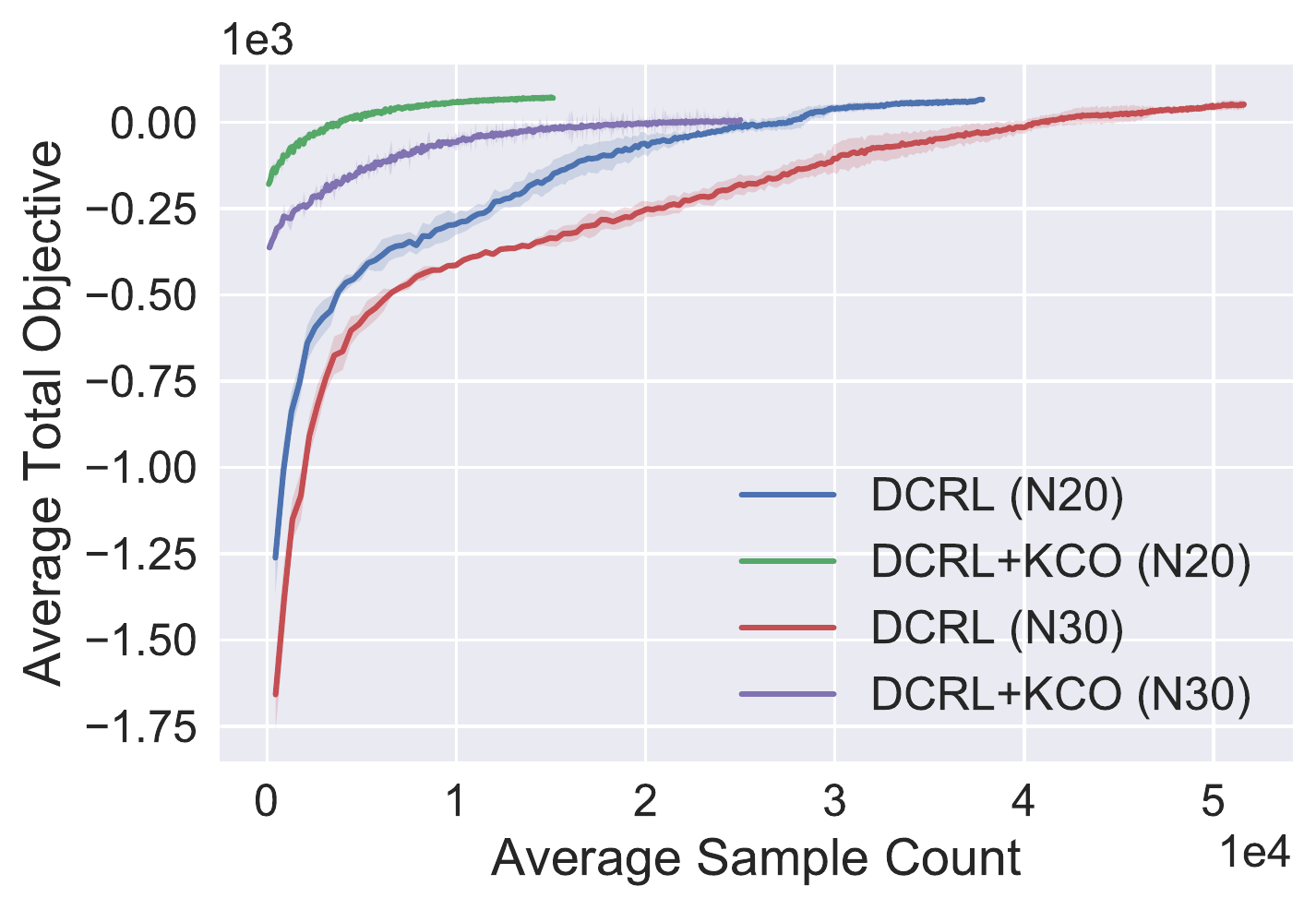}}
	\vskip -10pt
	\caption{\small Sample efficiency comparison between DCRL+KCO and DCRL on open grids (N\# denotes number of agents)}
	\label{fig:DCRL_open}
	\vskip -10pt
\end{figure*}  

\noindent{\textbf{Open Grid Maps: }}We next evaluate the integration of our knowledge-based framework with policy gradient and Q-learning based approaches. We combine our framework with DCRL~\cite{Ling0K20} and MAPQN~\cite{Fu2019} on several open grid maps with varying number of agents. DCRL is a policy gradient based algorithm, and MAPQN is a Q-learning based algorithm. We follow the same MAPF model as~\cite{Ling0K20} where each node has its own capacity (maximum number of agents that can be accommodated), and agents can take multiple time steps to move between two contiguous nodes. The total objective is to minimize sum of costs (SOC) of all agents combined with penalties for congestion. More details on the experiments, the neural network structure and the hyperparameters are noted in the supplementary material.% This set of experiments was run on a Linux machine with \textcolor{red}{(CPU, RAM???)}

The environment setting is varying from 4x4, 2 agents up to 10x10, 30 agents. We generated 10 instances for each setting. In each setting, we follow~\cite{Ling0K20} to randomly select the sources and destinations and to specify the capacity of each node. We also specify the min and max time ($t_{min}$, $t_{max}$) to move between two nodes. We run for each instance three times, and we terminate the runs either after 500 iterations or 10 hours. Each episode has a maximum length of 500 steps. For each instance, we choose the run with the best performance. We compute the total objective averaged over all agents and the cumulative number of samples averaged over all agents during training. Finally, we plot the average total objective vs the average cumulative sample count over all instances.

Figure~\ref{fig:DCRL_open} shows the results comparing DCRL with Knowledge Compilation (DCRL+KCO) and DCRL on 4x4, 8x8, and 10x10 grids (plots for 4x4, 2 agents, 8x8, 6 agents, and 10x10, 10 agents are deferred to the supplementary). Although all agents are able to reach their respective destinations (no stranded agents) in both DCRL+KCO and DCRL, agents are trained to reach destinations cooperatively with significantly fewer samples in DCRL+KCO. It means that agents are exploring the environment more efficiently in DCRL+KCO than in DCRL especially during the initial few training episodes. This is also reflected in the plot as the average total objective in DCRL+KCO is significantly higher during initial training phase compared to DCRL. 

\begin{figure}[t]
	\centering
	\subfigure[4x4 grid]{\includegraphics[scale=0.285]{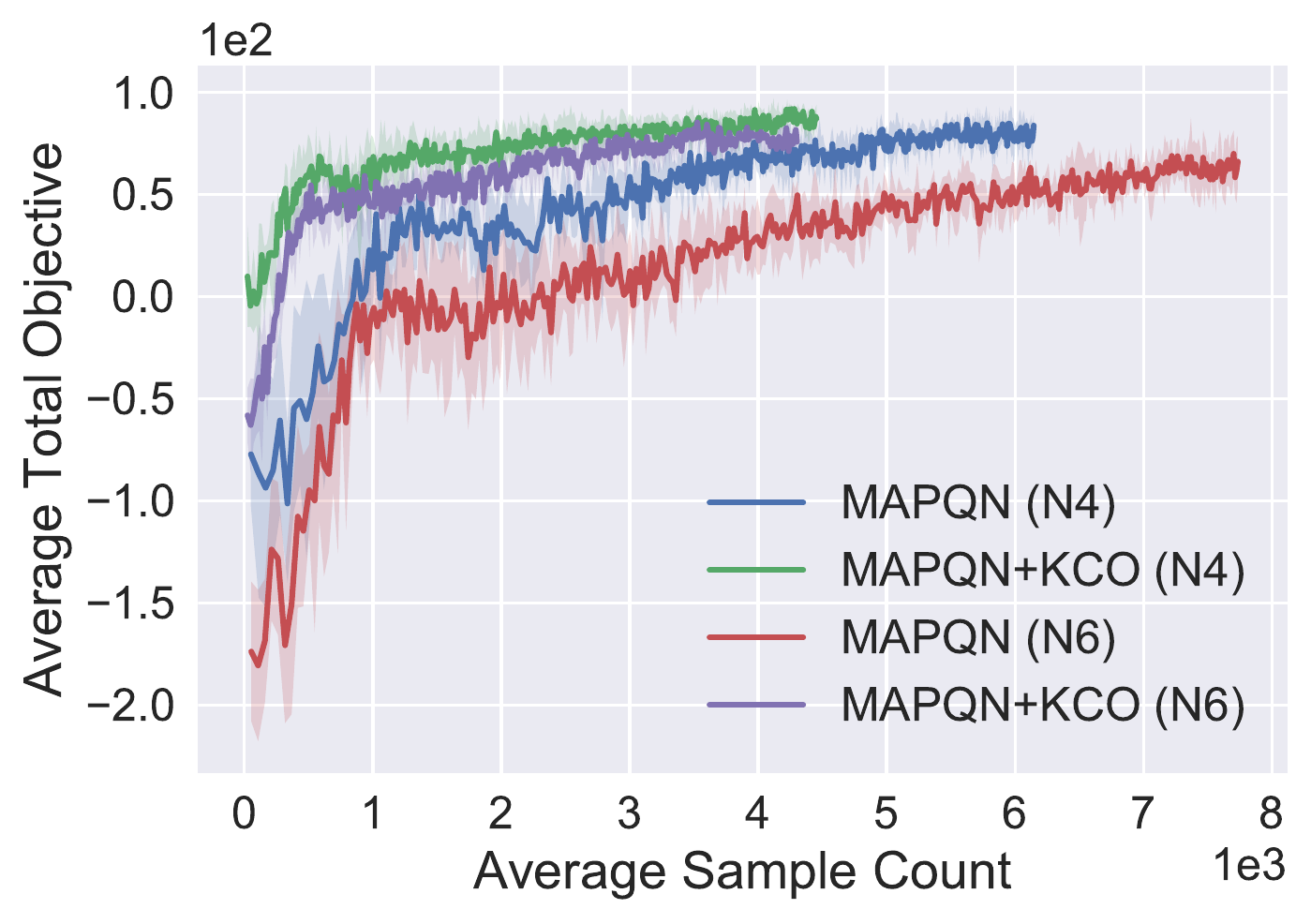}}
	\subfigure[8x8 grid]{\includegraphics[scale=0.285]{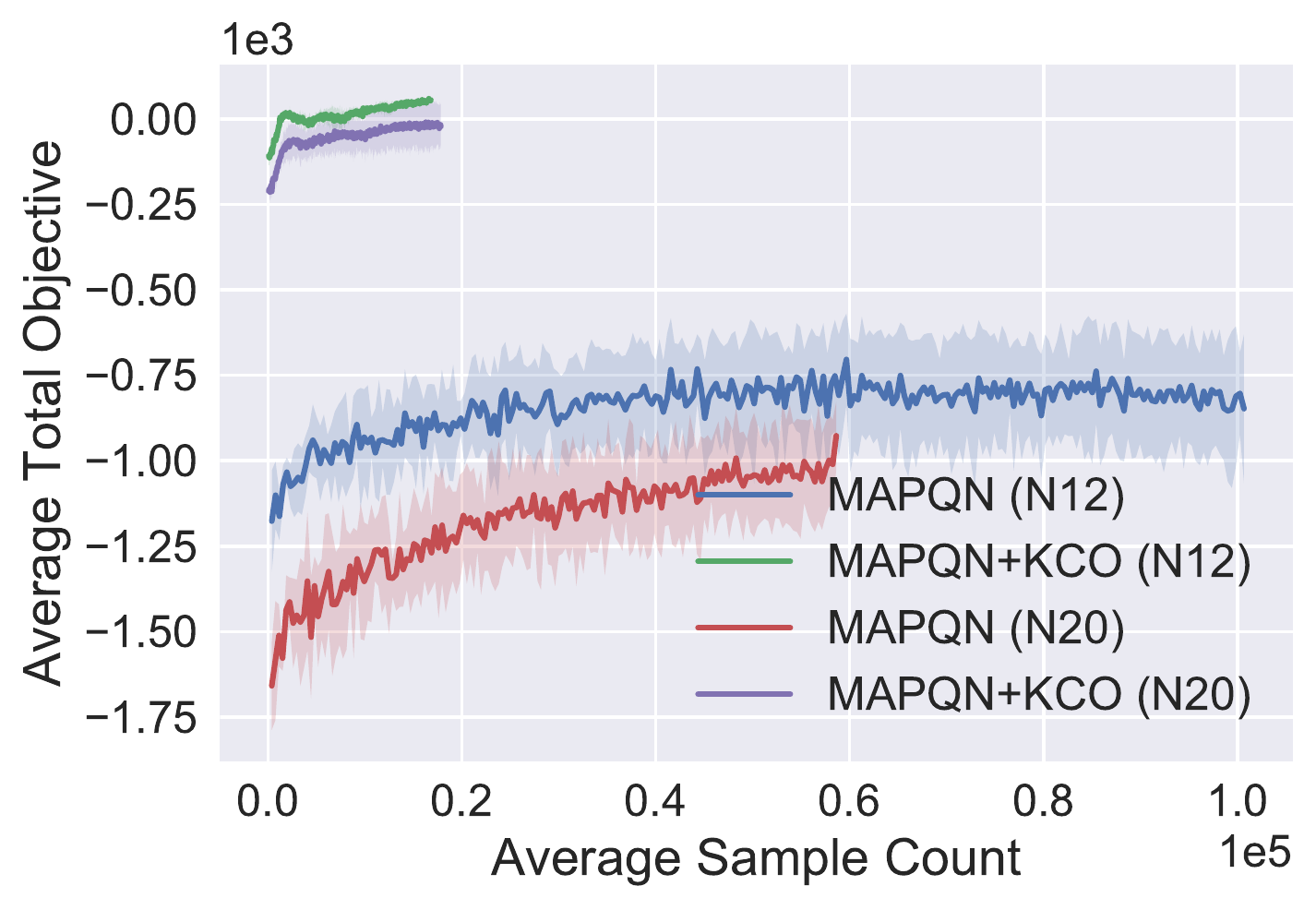}}
	\vskip -10pt
	\caption{\small Sample efficiency comparison between MAPQN+KCO and MAPQN on open grids (higher quality better)}
	\label{fig:MAPQN_open}
	\vskip -10pt
\end{figure}

Figure~\ref{fig:MAPQN_open} shows the comparison of sample efficiency between MAPQN+KCO and MAPQN on 4x4 and 8x8 grids. We did not evaluate MAPQN+KCO on 10x10 grid since MAPQN itself is not able to train a large number of agents on large grid maps (more details in~\cite{Ling0K20}). We observe that MAPQN+KCO converges faster and to a better quality than MAPQN especially on 8x8 grid. This is because several agents did not reach their destination within the episode cutoff in MAPQN, in contrast, all agents reach their destination in MAPQN+KCO.

%Table~\ref{tab:stranded_agents} shows the number of stranded agents on 8x8 grid. Since many agents were not able to reach their destinations, the average total objective after convergence in MAPQN is much lower than that in MAPQN+KCO. These results show that KCO can remarkably help find a better solution for all agents.
% These results clearly show that our KCO can remarkably help find better solutions for all agents. 

%\begin{table}[h]
%\centering
%\begin{tabular}{ |c|c|c|c| } 
% \hline
%	  & 12 agents & 20 agents \\ 
% \hline
%MAPQN+KCO & 0 & 0 \\ 
% \hline
%MAPQN & 9.8 & 12.4 \\ 
% \hline
%\end{tabular}
%	\vskip -5 pt
%	\caption{\small Number of stranded agents (average) on 8x8 grid by MAPQN+KCO and MAPQN (lower is better)}
%	\label{tab:stranded_agents}
%\end{table}

\begin{figure}[t]
	\centering
	\subfigure[DCRL+KCO]{\includegraphics[scale=0.285]{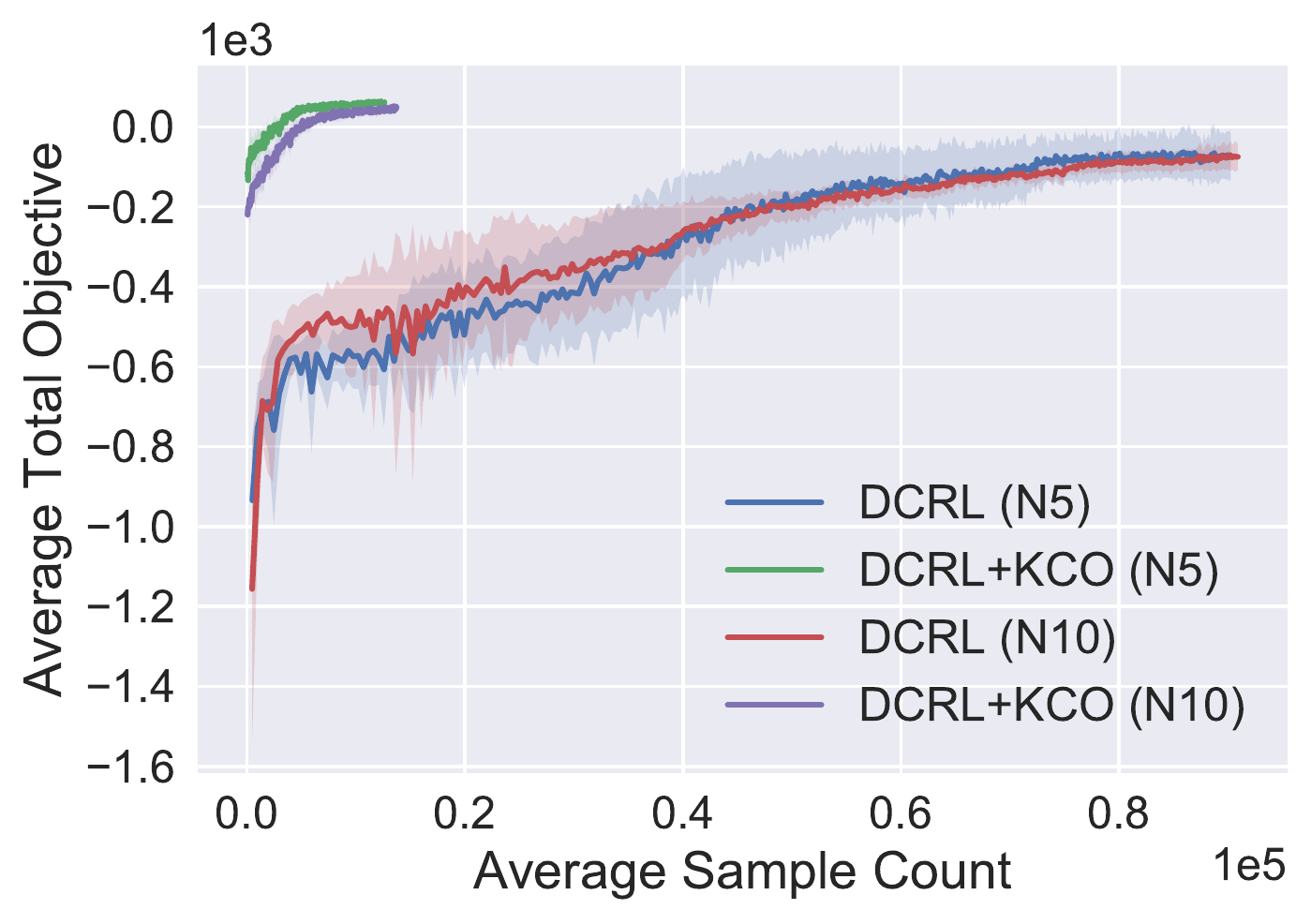}\label{fig:10_ob_DCRL}}
	\subfigure[MAPQN+KCO]{\includegraphics[scale=0.285]{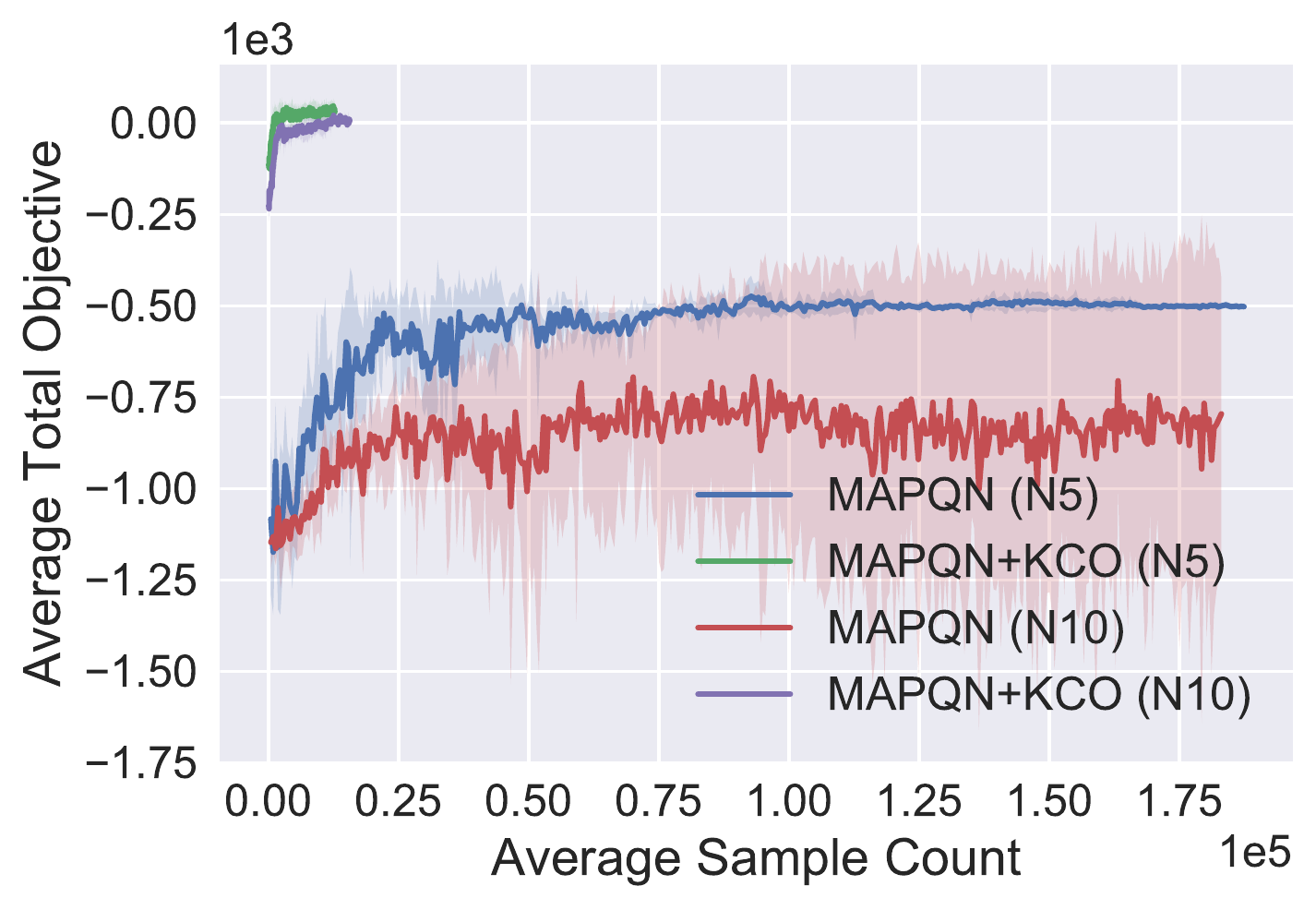}\label{fig:10_ob_MAPQN}}
	\vskip -10pt
	\caption{\small Sample efficiency results on 10x10 grid with obstacles}
	\label{fig:10_ob}
	\vskip -10pt
\end{figure}

\noindent{\textbf{Obstacles: }}We evaluate KCO with DCRL and MAPQN on a 10x10 obstacle map with varying number of agents (from 2 agents up to 10 agents). The obstacles are randomly generated with density 0.35. We generate 10 instances for this setting. For each instance, sources and destinations are randomly generated from the non-blocked nodes from the top and bottom rows (each source and destination pair is guaranteed to be reachable). Other parameters are specified in the same way as the above experiments. This set of experiments is quite challenging especially when there are several agents since they can go into dead ends easily while cooperating with each other to reduce the congestion level. 
Figure~\ref{fig:10_ob} clearly shows that DCRL and MAPQN can converge much faster with the integration of KCO and confirms that our approach is more sample efficient. Specifically for MAPQN, several agents did not reach their destinations (8.8 agents on average, for N10 case), whereas in MAPQN+KCO, all agents reached destination, which explains much better solution quality by MAPQN+KCO. %while also minimizing congestion. Therefore

%Table \ref{tab:stranded_agents_ob} shows that all agents reach their destinations in DCRL+KCO and MAPQN+KCO. 
%Therefore, the solution quality is better in DCRL+KCO and MAPQN+KCO than that in DCRL and MAPQN respectively.

%\begin{table}[h]
%\centering
%\resizebox{\linewidth}{!}{
%\begin{tabular}{ |p{1.5em}|c|c|c|c| } 
% \hline
%	  & DCRL+KCO & DCRL & MAPQN +KCO & MAPQN\\ 
% \hline
%N5 & 0 & 1 & 0 & 5\\ 
% \hline
%N10 & 0 & 2.2 & 0 & 8.8\\ 
% \hline
%\end{tabular}}
%	\vskip -5 pt
%	\caption{\small Number of stranded agents (average) on  10x10 grid with obstacles (N\# denotes number of agents) }
%	\label{tab:stranded_agents_ob}
%\end{table}

\begin{figure}[t]
	\centering
	\subfigure[2 agents]{\includegraphics[scale=0.285]{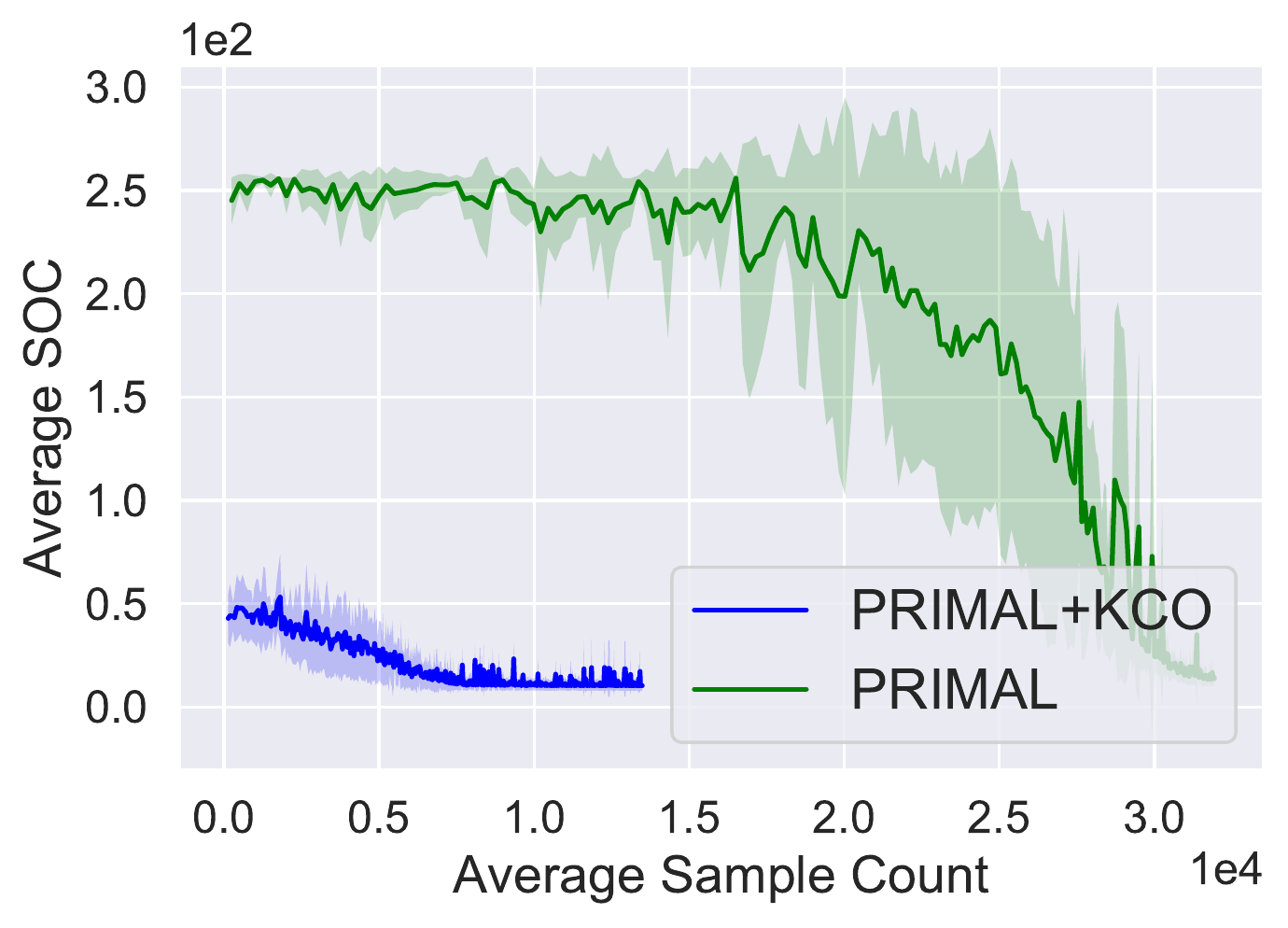}}
	\subfigure[4 agents]{\includegraphics[scale=0.285]{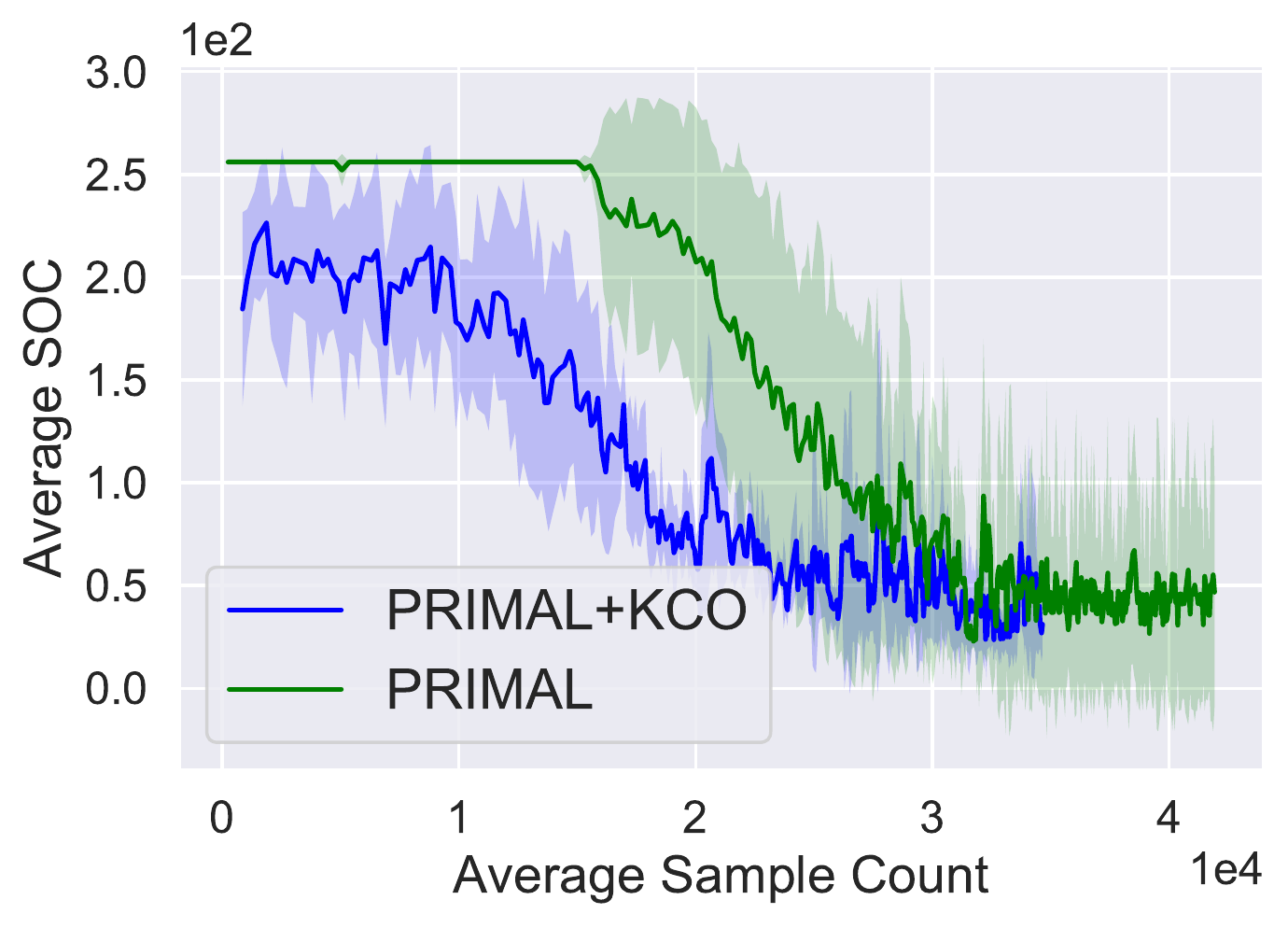}}
	\vskip -10pt
	\caption{\small Sample efficiency comparison on obstacle maps}
	\label{fig:primal_ob}
	\vskip -10pt
\end{figure}

We also evaluate KCO integrated with the PRIMAL framework~\cite{primal} which is based on asynchronous
advantage actor-critic or A3C~\cite{mnih2016asynchronous} combined with imitation learning. We test it on a 10x10 map with obstacles, keeping the density high (0.35). We generated 10 instances and tested with 2 and 4 agents. As noted in~\cite{primal}, high obstacle density is particularly problematic for PRIMAL.
%of the map with different sources and destinations for each agent and test it with 2 and 4 agents. PRIMAL does not work very well on small grids with large number of agents~\cite{primal}. Therefore, we didn't test it with more than 4 agents. 
Our results in Figure~\ref{fig:primal_ob} show that PRIMAL+KCO clearly outperforms PRIMAL in terms of sample efficiency. With 2 agents, the average SOC in PRIMAL is fluctuating around 250 during the initial episodes (maximum episode length is 256). However, the average SOC in PRIMAL+KCO is quite low during the initial training phase as expected (lower is better). With 4 agents, although the average SOC is quite high in both PRIMAL and PRIMAL+KCO during the initial episodes, the average SOC by PRIMAL+KCO is still lower than that by PRIMAL. The reason for the initial high average SOC is that the agents are trying to avoid collisions by taking a lot of $\nop$ actions because of the high density.
Overall, our framework is flexible enough to be integrated with different MARL approaches and consistently improve the performance.

\section{Conclusion}
We addressed the problem of cooperative multiagent pathfinding under uncertainty. Our work compiled static domain information such as underlying graph connectivity using propositional logic based decision making diagrams. We developed techniques to integrate such diagrams with deep RL algorithms such as Q-learning and policy gradient. Furthermore, to make simulation faster for RL, we developed an algorithm by analyzing the sub-context connectivity. We showed that the simulation speed of our algorithm is faster than the generic $\psdd$ method. We demonstrated the effectiveness of our approach both in terms of sample efficiency and solution quality on a number of instances.     

\onecolumn
\section*{Supplementary}
\section*{Appendix A}
\subsection*{Proof of Lemma 1}
\begin{proof}
Consider a $\psdd$ normalized for a right-linear vtree. A vtree is right-linear if each left child for each of its internal nodes is a leaf. Since primes are defined only for decision nodes, consider a $\psdd$ decision node $n$ normalized for a vtree node $v$. 
Using the definition of normalization, the primes primes $p_1,\mydots,p_k$ of $n$ are normalized for the left child of $v$. Since each left child in the vtree is a leaf (because the vtree is right-linear) which contains a single variable (let's say $X$), hence each of the primes $p_1,\mydots,p_k$ are literals $X$, $\neg X$ or the constant $\top$. \\
\textbf{Prime nodes encountered during sampling of a $\psdd$ encoding simple paths cannot be $\top$: }For a $\psdd$ normalized for a right linear vtree encoding simple paths between a source and a destination in a graph $G$, let $p_1,\mydots,p_k,s_k$ be the sampled $\psdd$ nodes (primes or sub) and let $X_1,\mydots,X_k,X_{k+1}$ be the corresponding literals (Lemma 1). Assume, on the contrary, that a prime $p_i=\top$. Because of the $\psdd$ semantics, the corresponding literal can be $X_i$ or $\neg X_i$. Only one of $X_i$ or $\neg X_i$ would form a simple path but not both. Hence our assumption was wrong and $p_i\ne\top$.\\
% We prove this by contradiction. Consider a sample $p_1,\mydots,p_l,s_l$ from such a $\psdd$ and assume $p_i=\top$ is normalized for a vtree node corresponding to the variable $X_i$ for some $i\in\{1,\mydots,l\}$. This sample is also an s-path and from Lemma 2, it also corresponds to a simple path in $G$ from $s$ to $d$. Since the s-path $p_1\land\mydots p_i\mydots\land s_l = p_1\land\mydots\top\mydots\land s_l = p_1\land\mydots\land s_l$, does not have $X_i$, it doesn't correspond to a simple path in $G$ which is a contradiction. Hence, $p_i\ne\top$.\\
\textbf{Example: }For example, in Figure 2(d) (main paper), all the primes in the $\psdd$ are literals and none of the primes are $\top$, since the $\psdd$ represents all simple paths between the nodes $n1$ and $n5$ in the graph $G$ in Figure 2(a) (main paper)
\end{proof}

\subsection*{Proof of Lemma 2}
\begin{proof}
\textbf{Mapping: }Let $\spset$ denote the set of all feasible s-paths in a $\psdd$ that encodes simple paths between a source $s$ and a destination $d$ in an undirected graph $G=(V,E)$. Also, let $\smset$ denote the set of all simple paths between $s$ and $d$ in $G$. Now consider the mapping $f(sp)=sm$, $\forall sp\in\spset$ and $\forall sm\in\smset$, which maps all elements ($p_i$ or $n$) in $sp$ such that we include the edge corresponding to the literal of the element in our path if the literal is positive and we don't include it if it's is negative. This is true because each prime is a literal corresponding to an edge in $G$ (Lemma 1).\\ 
\textbf{Example: }As an example, consider the $\psdd$ in fig 2(d) and an s-path $sp=A\land\neg B\land C\land\neg D\land E$ indicated by the red arrows. Now consider the mapping $f$ where the prime $A$ is mapped to the edge $(n1,n2)$, $B$ is mapped the edge $(n2,n3)$ etc. as shown in fig 2(a). Then $sp$ represents the path $A-C-E$ in $G$.\\
\textbf{f is one-to-one: }To show that $f$ is one-to-one, assume otherwise. Let $sp_1$ and $sp_2$ be two different feasible s-paths for which $f(sp_1)=f(sp_2)$. If $sp_1$ and $sp_2$ are different, there exists at least one element $x_1$ in $sp_1$ which is different from $x_2$ in $sp_2$ ($x_1, x_2$ are $p_i$ or $n$). But since $x_1$ and $x_2$ also correspond to edges, $f(sp_1)$ and $f(sp_2)$ represent two different simple paths in $G$, which is false. Hence $sp_1 = sp_2$ and $f$ is one-to-one.\\
\textbf{Note: }We can also show the other way, i.e., the set of all paths from source $s$ to destination $d$ in $G$ can be mapped to s-paths in the set $\spset$. Consider a simple path from $s$ to $d$. Now, start from the root of the $\psdd$ and map edge $e$ to its corresponding literal $X_e$ if it is present in the simple path and if $e$ is not in the simple path, map it to $\neg X_e$ and keep going down the $\psdd$ until the last node (prime or sub). This forms an s-path and is feasible because if it was not, then one of the false sub would have made everything below it false (see proof of step 3 of the procedure). This mapping is one-to-one as well and can be proved in a similar manner as described above.
\end{proof}

\subsection*{Proof of step 3 of the procedure}
\begin{proof}
$sc^{\star}$ can be extended to a feasible s-path $sp\in\spset$ such that $sp$ contains $sc^{\star}$ (or $sc^{\star}\land sp = sp$).\\
To show that $sc^{\star}$ can be extended to a feasible s-path, we first start from the $\psdd$ node for which $sc^{\star}$ is defined and go down the $\psdd$ till the deepest node (prime or sub) and selecting the primes (or the sub) encountered and constructing an s-path $sp$. We show $sp$ is feasible by contradiction. Assume that there's no feasible s-path that can be constructed from $sc^{\star}$. This implies that all subs encountered in the path from $s^{\star}$ to the deepest node in the $\psdd$ are false. This, in turn, implies that the sub of the corresponding prime for which $sc^{\star}$ is defined is false too. But this cannot be true since $sc^{\star}$ is a feasible sub-context. Hence, there is at least one s-path $sp$ to which $sc^{\star}$ can be extended, i.e., $sc^{\star}\land sp=sp$.\\ 
\textbf{Example: }Suppose $sc^{\star}$ is the sub-context for the node $C$ and is given by $A\land\neg B\land C$. If we go down the $\psdd$ and select the literals encountered, i.e., $\neg D, E$, we can construct a feasible s-path $A\land\neg B\land C\land\neg D\land E$.\\
\textbf{Note: }In the procedure, if $\tilde{X}$ represents a sub, which only happens if $\tilde{X}$ is the deepest in the vtree, we check if a sub-context $sc\in\sset{\tilde{X}}$ contains all the positive literals $\{X_{e_1},\mydots,X_{e_k}\}$ (i.e. we do not check for $X_e$).
\end{proof}

\subsection*{Proof of step 4 of the procedure}
\begin{proof}
If $sc^{\star}$ does not exist then a feasible s-path cannot be found containing all the literals $\{X_{e_1},\mydots,X_{e_k},X_e\}$.\\
We can easily show this by contradiction. Assume, on the contrary, that if $sc^{\star}$ does not exit then there exists a feasible s-path $sp$ exists containing all the literals $\{X_{e_1},\mydots,X_{e_k},X_e\}$. Since $sc^{\star}$ does not exist, the sub-context $sc$ that we are extending to $sp$ does not contain at least one of the variables in $\{X_{e_1},\mydots,X_{e_k},X_e\}$. But this implies $sp$ is not a valid s-path. Therefore, if $sc^{\star}$ does not exist then a feasible s-path cannot be found. 
% We do this because when $\tilde{X}$ represents a sub, it's not present in its own sub-context and if we select its corresponding prime then we select the sub too.
\end{proof}

\subsection*{Example of the procedure}
Consider the graph in Figure 2(a) (main paper) with source $s=n1$ and destination $d=n5$. The corresponding $\psdd$ and vtree is given in Figure 2(c) (main paper) and Figure 2(d) (main paper). Let the partial path be $\pth$ with edges $\{(n1,n2),(n2,n3)\}$ or their corresponding Boolean variables $\{A, B\}$. Now we want to find if the edge $e=(n3,n4)$ can be selected, i.e., if $Pr(X_e=D|A,B)>0$. First we find $\tilde{X}$, which turns out to be the literal $D$. We also compute $\sset(D)=\{(A\land B\land\neg C\land D)\}$. We can clearly see that $sc^{\star}=(A\land B\land\neg C\land D)$ contains all the literals in the set $\{A, B\}$. Now we see that $sc^{\star}$ can be extended to an $\spath$ $sp=(A\land B\land\neg C\land D\land E)$ since $sc^{\star}\land sp = sp$.

\subsection*{Optimization of step 2}
We now present how we optimize step number 2 by pre-processing and pruning of sub-contexts. Assume the partial path is $\pth = \{e_1, \mydots, e_k\}$ in the graph $G$. Let $\tilde{X} \in \{X_{e_1}, \mydots, X_{e_k}\}$ be the deepest variable in the vtree order. Let $\sset^*(\tilde{X})$ be the set where each element $sc'$ in the set satisfies the constraint $sc' \!\land\! X \!=\! sc',\ \forall X \!\in\! \{X_{e_1}, \mydots, X_{e_k}\}$. Now we look at one possible edge $e$ that the agent can traverse next given the partial path. Assume the corresponding Boolean variable is $X_e$. Step 2 could be executed as follows:
\begin{itemize}
\item Case 1: If $\tilde{X}$ is deeper than $X_e$, given $sc'\!\in\!\sset^*(\tilde{X})$, we check  $\exists\ sc\!\in\!\sset(X_e)$ s.t. $sc' \!\land\! sc \!=\! sc'$. 
\item Case 2: If $X_e$ is deeper than $\tilde{X}$, given $sc'\!\in\!\sset^*(\tilde{X})$, we check  $\exists\ sc\!\in\!\sset(X_e)$ s.t. $sc \!\land\! sc' \!=\! sc$. 
\end{itemize}
Intuitively, if we have $sc' \!\land\! sc \!=\! sc'$ or $sc \!\land\! sc' \!=\! sc$, then there is a path in $\psdd$ connecting one $\psdd$ prime node whose sub-context is $sc$ and the other $\psdd$ prime node whose sub-context is $sc'$. If there exists at least one sub-context $sc$, then edge $e$ is the edge that the agent can traverse next given the partial path $\pth = \{e_1, \mydots, e_k\}$. When the partial path actually becomes $\pth = \{e_1, \mydots, e_k, e\}$, we will update $\sset^*(\tilde{X})$ or create $\sset^*(X_e)$ (we will describe later). We now prove the correctness of this optimization of step 2. 
\begin{itemize}
\item Case 1: For each $sc'\!\in\!\sset^*(\tilde{X})$, we have $sc' \!\land\! X \!=\! sc',\ \forall X \!\in\! \{X_{e_1}, \mydots, X_{e_k}\}$. If there exist a sub-context $sc\!\in\!\sset(X_e)$ and a sub-context $sc'\!\in\!\sset(\tilde{X})$ such that $sc' \!\land\! sc \!=\! sc'$, then we will also have $sc' \!\land\! X_e \!=\! sc'$. Since $sc'$ is the sub-context of the variable deepest in the vtree order among $\{X_{e_1}, \mydots, X_{e_k}, X_e\}$, it can be extended to a feasible s-path $sp\in \spset$ such that $sp$ contains $sc'$ (proved earlier).
\item Case 2: For each $sc'\!\in\!\sset^*(\tilde{X})$, we have $sc' \!\land\! X \!=\! sc',\ \forall X \!\in\! \{X_{e_1}, \mydots, X_{e_k}\}$. If there exist a sub-context $sc\!\in\!\sset(X_e)$ and a sub-context $sc'\!\in\!\sset(\tilde{X})$ such that $sc \!\land\! sc' \!=\! sc$, then we will have $sc \!\land\! X \!=\! sc,\ \forall X \!\in\! \{X_{e_1}, \mydots, X_{e_k}, X_e\}$. Since $sc$ is the sub-context of the variable deepest in the vtree order among $\{X_{e_1}, \mydots, X_{e_k}, X_e\}$, it can be extended to a feasible s-path $sp\in \spset$ such that $sp$ contains $sc$ (proved earlier).
\end{itemize}

%thus the time complexity of the  $\nz$ inference for edge $e$ is $O(1)$.
\noindent{Evaluating} $sc' \!\land\! sc \!=\! sc'$ or $sc \!\land\! sc' \!=\! sc$ can be done in advance which is the pre-processing step to check connectivity of two sub-contexts. Now we present Algorithm~\ref{algo:pre-processing} to describe this pre-processing. Intuitively, we give each sub-context $sc\!\in\!\sset(X)$ a unique ID, and we store the IDs of any two sub-contexts $sc\!\in\!\sset(X)$ and $sc'\!\in\!\sset(X')$ that are connected in a hash table. Algorithm~\ref{algo:inference} describes the $\nz$ inference for a possible edge $e$ given the partial path $\pth = \{e_1, \mydots, e_k\}$ in graph $G$ by using the $connectivity$ hash table returned from Algorithm~\ref{algo:pre-processing}. When updating $\sset^*(\tilde{X})$, we prune sub-context $sc'$ whose $ID$ is not in the $IDsToUpdate$; When creating $\sset^*(X_e)$, 
we will add $sc$ from $\sset(X_e)$ whose ID is in $IDsToUpdate$.

\vskip 10pt
\begin{algorithm}[H]
 \textbf{Input}: $\sset(X),\ \forall X \in \{X_1, \ldots, X_n\}$ (set of vtree variables);\\
 
  $connectivity = HashTable()$ \\
  \For{$X$ in $\{X_1, \ldots, X_n\}$}{
  
  		\For{$X'$ in $\{X_1, \ldots, X_n\}\setminus\{X\}$}{

  \eIf{$X'$ is deeper than $X$ in vtree}
  	{
  	$connectivity[(X, X')] = HashTable()$			\\ 
		\For{$sc,\ scID$ in enumerate($sset(X))$}
		{	
			$connectivity[(X, X')][scID] = List()$			\\ 
			\For{$sc',\ sc'ID$ in enumerate($sset(X'))$}
			{
				\If{$sc' \!\land\! sc \!=\! sc'$}
				{
					add $sc'ID$ to $connectivity[(X, X')][scID]$
				}
			}
		
		}	 
	          
    }
	{
  	$connectivity[(X, X')] = HashTable()$			\\ 
		\For{$sc,\ scID$ in enumerate($sset(X))$}
		{	
			$connectivity[(X, X')][scID] = List()$			\\ 
			\For{$sc',\ sc'ID$ in enumerate($sset(X'))$}
			{
				\If{$sc \!\land\! sc' \!=\! sc$}
				{
					add $sc'ID$ to $connectivity[(X, X')][scID]$
				}
			}
		
		}

	}  
 
}
}
\textbf{Output}: $connectivity$\\
\caption{Pre-processing of sub-contexts}
\label{algo:pre-processing}
\end{algorithm}

\vskip 10pt
\begin{algorithm}[H]
 \textbf{Input}: $connectivity$,  $\pth = \{e_1, \mydots, e_k\}$, $\sset^*(\tilde{X})$, a possible edge $e$\\
   	  
  \eIf{$\tilde{X}$ is deeper than $X_e$}
  	{
				
		\For{$sc',\ sc'ID$ in enumerate($\sset^*(\tilde{X})$)}  
			{	
				$IDsToUpdate = List()$ \\
				
				\If{$connectivity[(\tilde{X}, X_e)][sc'ID] \neq \emptyset$}
				{
					add $sc'ID$ to $IDsToUpdate$\\
				} 
			} 
    }
	{
 		\For{$sc',\ sc'ID$ in enumerate($\sset^*(\tilde{X})$)}  
			{
				$IDsToUpdate = List()$\\
				\If{$connectivity[(\tilde{X}, X_e)][sc'ID] \neq \emptyset$}
				{
					add $connectivity[(\tilde{X}, X_e)][sc'ID]$ to $IDsToUpdate$\\
				} 
			} 
	}  
  \If{$IDsToUpdate \neq \emptyset$}
  {
  		$e$ is the edge that the agent can traverse next.\\
  }

	\If{$\pth = \{e_1, \mydots, e_k, e\}$}
	{
		\eIf{$\tilde{X}$ is deeper than $X_e$}
		{
		 	Update $\sset^*(\tilde{X})$ 
		}
		{
			Create $\sset^*(X_e)$ 
		}
	}
\caption{$\nz$ inference}
\label{algo:inference}
\end{algorithm}

\pagebreak
\subsection*{Route distribution and map partitioning}
% Given an undirected graph $G$ and a set of Boolean variables $\mathbf{X}=\{X_1,\ldots,X_n\}$ each corresponding to an edge in $G$, then $Pr(\mathbf{X})$ is a distribution on all the edges. By putting constraints on $\mathbf{X}$, one can define distributions on these constrained spaces. We define these constraints using propositional logic and represent them as SDDs. Also, we represent the resulting distributions using PSDDs. E.g., if $\alpha$ is the Boolean formula representing the constraint that edges represent a path in $G$ then $Pr(\mathbf{X})$ represent a distribution on paths, i.e., $Pr(\mathbf{X})$ is a distribution representing all paths in $G$ iff $Pr(\mathbf{x})=0$ if $\mathbf{x}\not\models\alpha$. Similarly, one can define distribution on simple paths. 

\begin{figure}[!ht]
	\centering
	\subfigure[]{\includegraphics[scale=0.25]{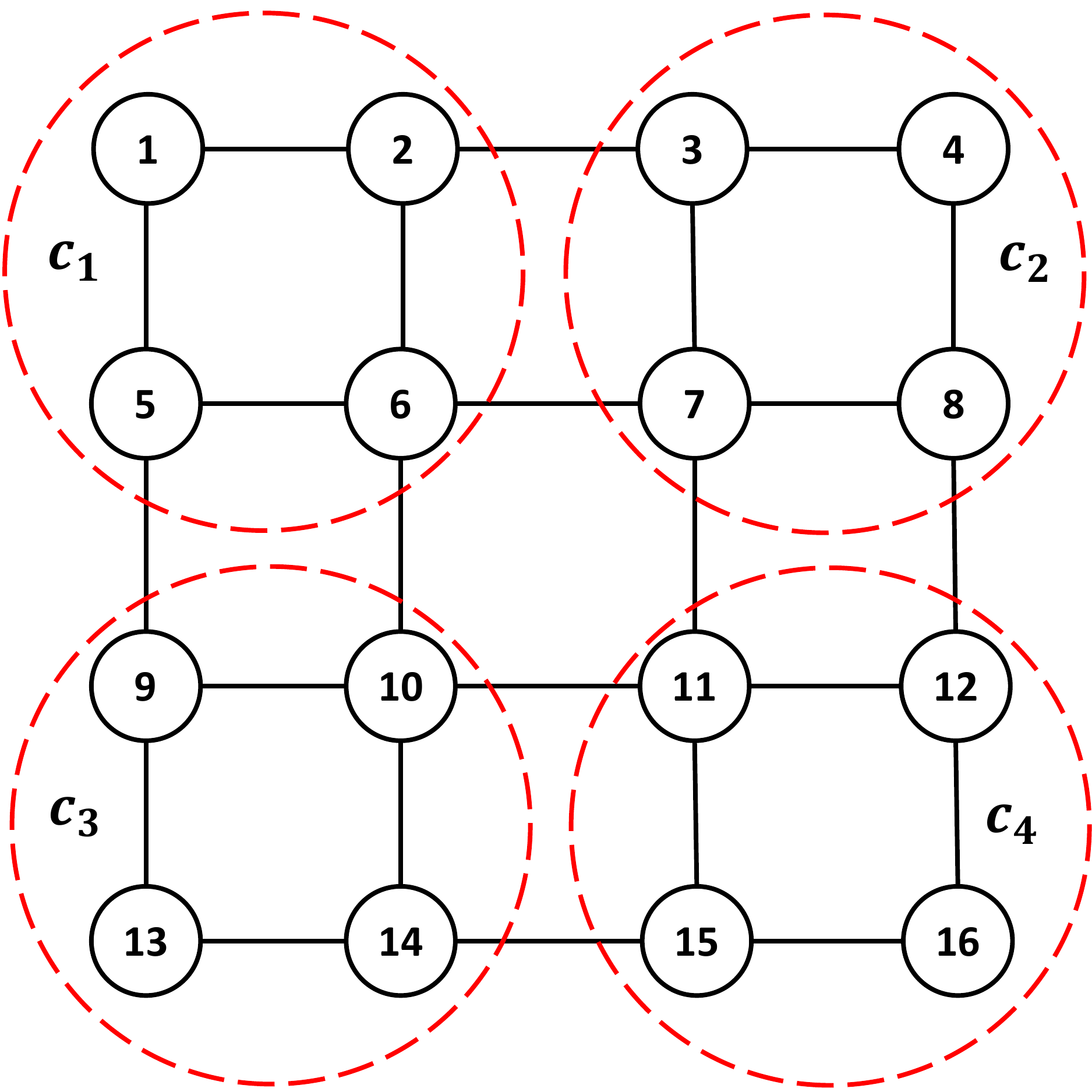}}
	\subfigure[]{\includegraphics[scale=0.25]{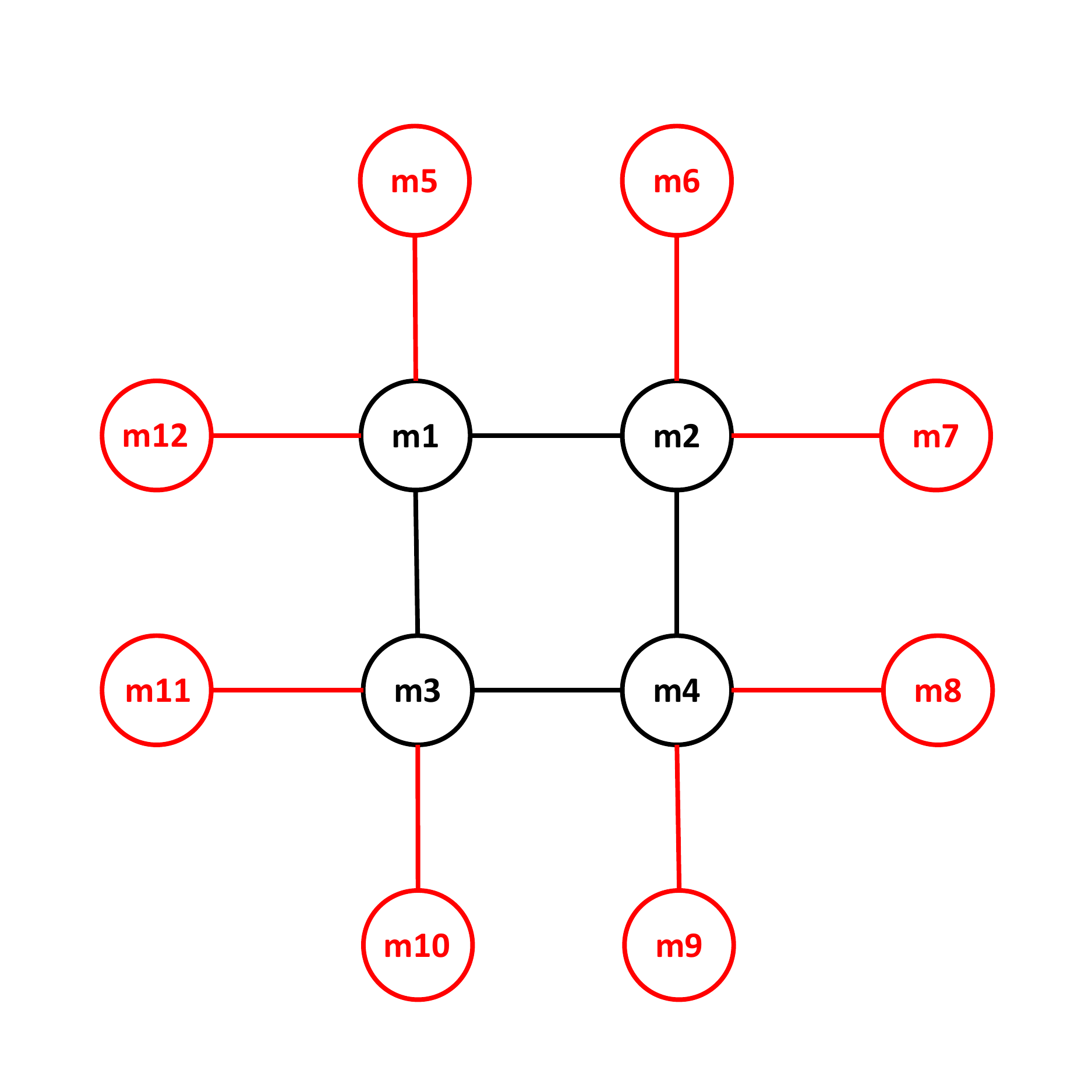}}
	\vskip -10pt
	\caption{\small (a)A 4x4 grid map partitioned into regions (or clusters) $c_1,\mydots,c_4$. $c_1,\mydots,c_4$ form a 2x2 grid map $G_p$ (b) Inside of a cluster, e.g., $c_1$. The black edges are the inernal edges and the red edges are the external edges.}
	\label{fig:map_partition}
	\vskip -10pt
\end{figure}

% Given a map as an undirected graph $G$ and a set of Boolean variables $\mathbf{X}=\{X_1,\ldots,X_n\}$ each corresponding to an edge in $G$, one can define constraints on $\mathbf{X}$ using propositional logic and represent them as an SDD. Furthermore, one can also define a distribution $Pr(\mathbf{X})$ on this constraint space. E.g., $Pr(\mathbf{X})$ can represent distribution over simple paths in $G$.
\noindent{T}o partition a map represented as an undirected graph $G=(V,E)$, we partition its nodes $V$ into regions or clusters $c_1,\mydots,c_m$, with each cluster $c_i$ having {\it internal} and {\it external} (that cross into $c_i$) edges. On these clusters, we induce a graph $G_p$ with $c_1,\mydots,c_m$ as nodes. We then define constraints on $\mathbf{X}$ using $G$ and $G_p$ that paths that are simple in $G_p$ are also simple w.r.t $G$ and induce a distribution $Pr(\mathbf{X})$ over them. More concretely, paths cannot enter a region twice and they also cannot not visit any nodes inside the clusters twice. We represent all the simple paths \textit{inside} the clusters $c_1,\mydots,c_m$ and also \textit{across} the clusters as $\psdd$s. This is a hierarchical representation of paths in which we have two levels of hierarchy, one for across the clusters and another for inside the clusters.\\
\textbf{Example: }Consider Figure~\ref{fig:map_partition}(a), where a 4x4 grid map is partitioned into clusters $c_1,\mydots,c_4$ and the graph $G_p$ is formed from these clusters as nodes. Figure~\ref{fig:map_partition}(b) represents inside of a cluster which is a 2x2 grid map. The black edges are the internal edges and the red edges are the external ones. We construct $\psdd$s between all the red nodes inside the cluster and also a $\psdd$ for the 2x2 grid map formed by $c_1,\mydots,c_4$. Let's say Figure~\ref{fig:map_partition}(b) represents cluster $c_1$, i.e., node 1 is mapped to $m1$, 2 is mapped to $m2$ and so on. Similarly, edge $(m2,m7)$ is mapped to the edge $(2,3)$, $(m4,m8)$ to $(6,7)$ and so on. Now, to sample a path that starts from node 1, we start from $m12$ (or $m5$) to enter $c_1$. We keep sampling until we encounter an external edge. If, for example, we encounter the edge $(m4,m8)$, we traverse the edgeg $(6,7)$ in the 4x4 grid and move to the cluster $c_2$ and keep sampling until we reach the destination. (We discard $(m2,m6)$ for $c_1$ because it is not mapped to any of the edge in the 4x4 grid). 

% \section*{Appendix B: Extensions}
% The kind of knowledge-compilation framework we presented can be extended to settings where an agent is required to visit some landmarks before reaching the destination. We can easily include the landmarks in the evidence set for calculating the next set of valid edges (e.g., computing {\small $Pr((v_{\pth}, v') | \pth, \textit{landmarkEvidence})$}). Furthermore, this framework can also be used in cases where the underlying graph connectivity is dynamic; e.g.,  in scenarios where edges are dynamically getting blocked over time. Any observation about blocked edges at a time can become the evidence, and by conditioning on this evidence, the agent can rule out routes via such blocked edges.
% This framework's generalizability and flexibility make it a promising approach in combining domain knowledge with models for RL, pathfinding, and other areas.

\section*{Appendix B: Empirical Evaluation}
\noindent{To} represent $\psdd$ and $\sdd$ in our experiments, we use the \texttt{GRAPHILLION}~\citep{inoue2016graphillion} package to first construct a ZDD and then convert it to $\sdd$~\citep{nishino2016zero,nishino2017compiling}. We also use the \texttt{PySDD}~\citep{darwiche2018recent} package for constructing $\sdd$s and \texttt{PyPSDD}\footnote{https://github.com/art-ai/pypsdd} package for constructing and doing inference on $\psdd$.

\noindent{\textbf{Simulation Speed: }}We evaluated the sampling speed of the $\psdd$ inference method for computing conditional probabilities~\citep{Kisa} and our approach based on sub-context connectivity analysis for $\nz$ inference (SCANZ) in open grid maps of different sizes 3x3, 4x4, 5x5, and 10x10. The experiments were performed on a single desktop machine with an Intel i7-8700 CPU and 32GB RAM (a 64 cores CPU and 256GB RAM machine for 10x10 grid). For each map, the source and destination are the top right node and bottom left node respectively. We randomly generate 10,000 paths given the source and destination pairs using both SCANZ and $\psdd$ conditional probabilities and calculate the running time for the entire path simulation. Table~\ref{simulation_speed} shows that SCANZ is more than an order of magnitude faster than $\psdd$ inference. \footnote{To test $\psdd$ inference on 10x10, we use the code from here:
https://github.com/hahaXD/hierarchical\_map\_compiler, 
which is based on~\citep{choi2017tractability,shen2019structured}}

\begin{figure}[h]
	\centering
	\subfigure[]{\includegraphics[scale=0.2]{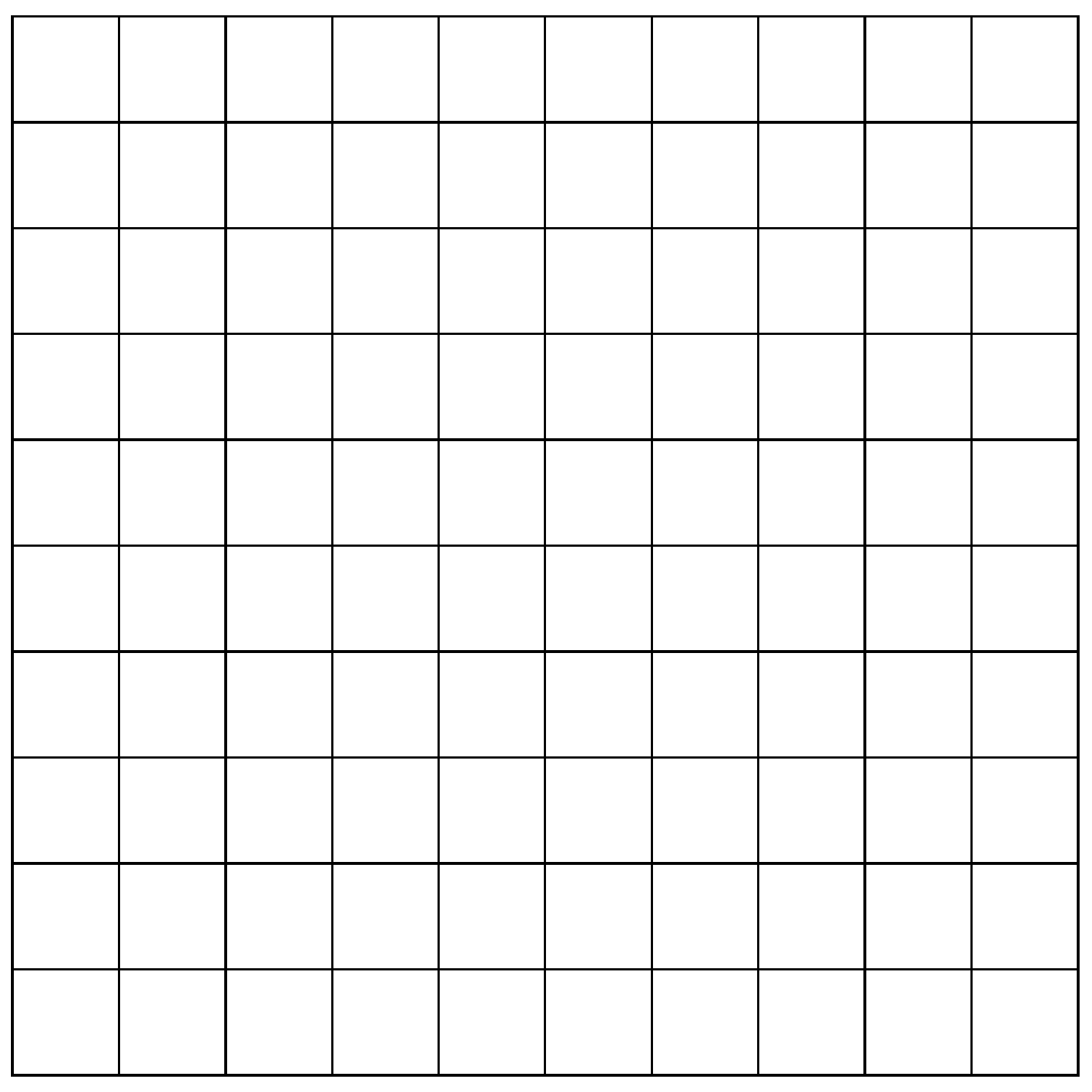}}
	\hspace{5em}
	\subfigure[]{\includegraphics[scale=0.2]{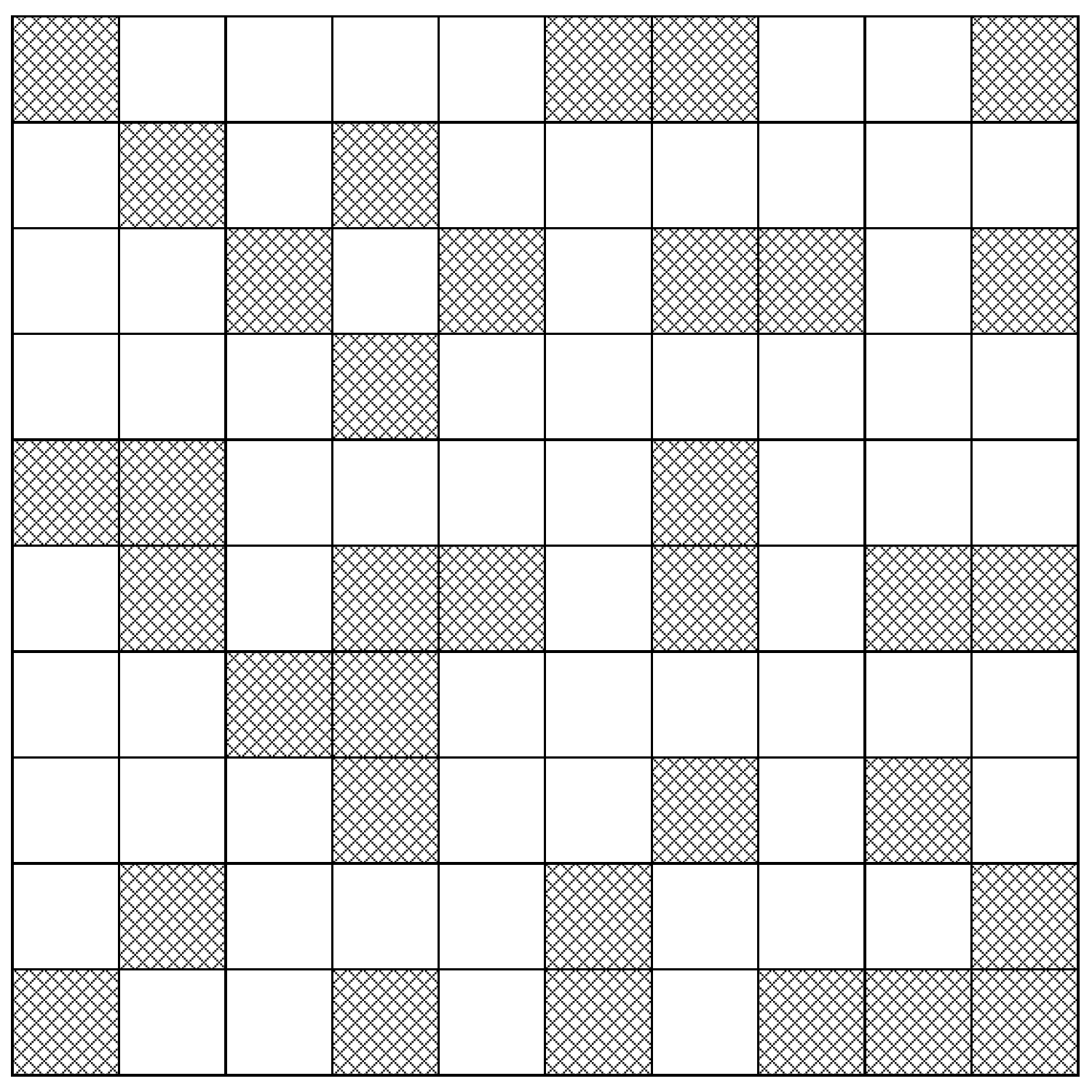}}
	\vskip -10pt
	\caption{\small (a) 10x10 open grid map; (b) 10x10 grid with obstacles (0.35 obstacle density, dark nodes are blocked) }
	\label{fig:map}
	\vskip -10pt
\end{figure}

\begin{table}[H]
\centering
\begin{tabular}{ |c|c c c|c| } 
 \hline
 \multirow{2}{5em}{\vspace{-0.5em} \textbf{Approach }} & \multicolumn{3}{|c|}{\textbf{nonhierarchical}} & \multicolumn{1}{|c|}{\textbf{hierarchical}} \\ \cline{2-5}
        & \multicolumn{1}{|c|}{3x3} & \multicolumn{1}{|c|}{4x4} & \multicolumn{1}{|c|}{5x5} & \multicolumn{1}{|c|}{10x10} \\		
 \hline
SCANZ & 1.84 & 3.86 & 19.82 & 407.95\\ 
 \hline
$\psdd$ inference & 26.55 & 158.41 & 979.71 & 402665.98\\ 
 \hline
\end{tabular}
	%\vskip -10pt
	\caption{\small Simulation speed comparison (in seconds)}
	
	\label{simulation_speed}
\end{table}
\vskip -10pt
\noindent{\textbf{Experimental Settings:}} To compare our approach with DCRL, we follow the same settings in ~\citep{Ling0K20}. For each grid map, sources and destinations are the top and bottom rows. For each agent, we randomly select its source and destination from the top and bottom row. The capacity of each node is sampled uniformly from [1, 2] for 4x4 grid, [1, 3] for 8x8 grid, and [1, 4] for 10x10 grid. For 10x10 grid with obstacles (as shown in Figure~\ref{fig:map}(b)), the capacity of each node is sampled uniformly from [1, 2] for 2 agents, [1, 3] for 5 agents, and [1, 4] for 10 agents. The $t_{min}$, $t_{max}$ for moving between two contiguous zones are 1, 5 respectively. We used the same 10x10 grip with obstacle map for evaluating PRIMAL+KCO and PRIMAL. The locations of obstacles are fixed. We generated 10 instances for 2 agents, 5 agents, and 10 agents respectively. For each instance, the source and destination for an agent are randomly selected from the non-blocked nodes. We run each instance for three times, and select the run with the best performance.

\noindent{\textbf{Hyperparameters and Neural Network Architecture: }}To compare our DCRL+KCO and MAPQN+KCO with DCRL and MAPQN respectively, we use the same  hyperparameters as in ~\citep{Ling0K20}. The neural network architectures are also the same except for the last $softmax$ layer in DCRL code. Instead, we use a customized layer to generate a probability distribution over all actions according to Equation (2) in the main paper. For comparing the PRIMAL framework with our approach, we use the same neural network architecture as in ~\citep{primal}. We also keep all the hyperparameters same when evaluating PRIMAL+KCO. We make a small change to get the final set of valid actions that the agent can take: we take intersection of the set of valid actions given by the PRIMAL environment ($validActions$) with the action set obtained by doing $\nz$ inference ($psddActions$). Concretely, the final set of valid actions that an agent can take is $validActions \cap psddActions$.

% In PRIMAL,  \footnote{https://github.com/gsartoretti/distributedRL\textunderscore MAPF},

\begin{figure}[H]
	\centering
	\subfigure[4x4 grid]{\includegraphics[scale=0.4]{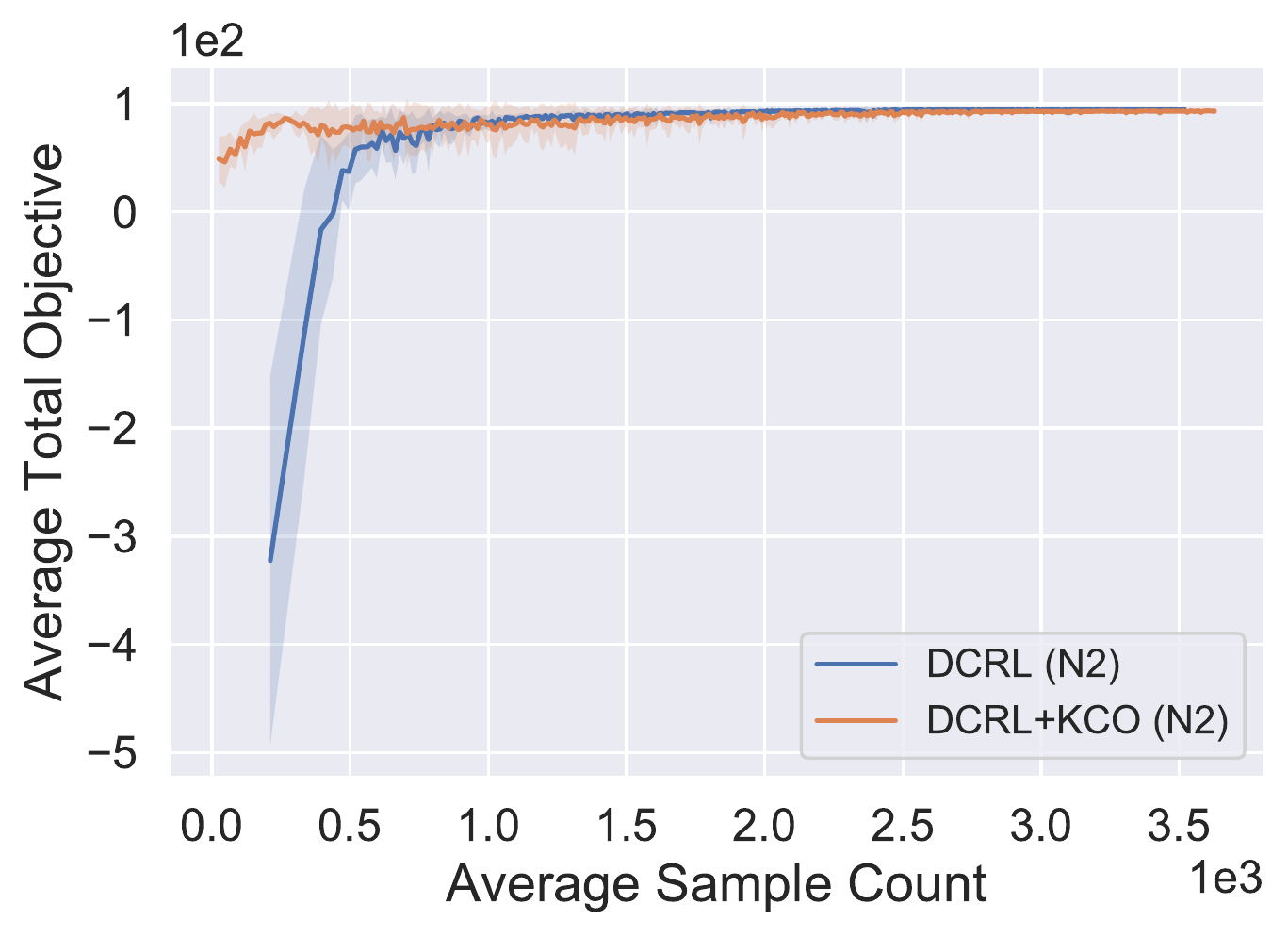}}
	\subfigure[8x8 grid]{\includegraphics[scale=0.4]{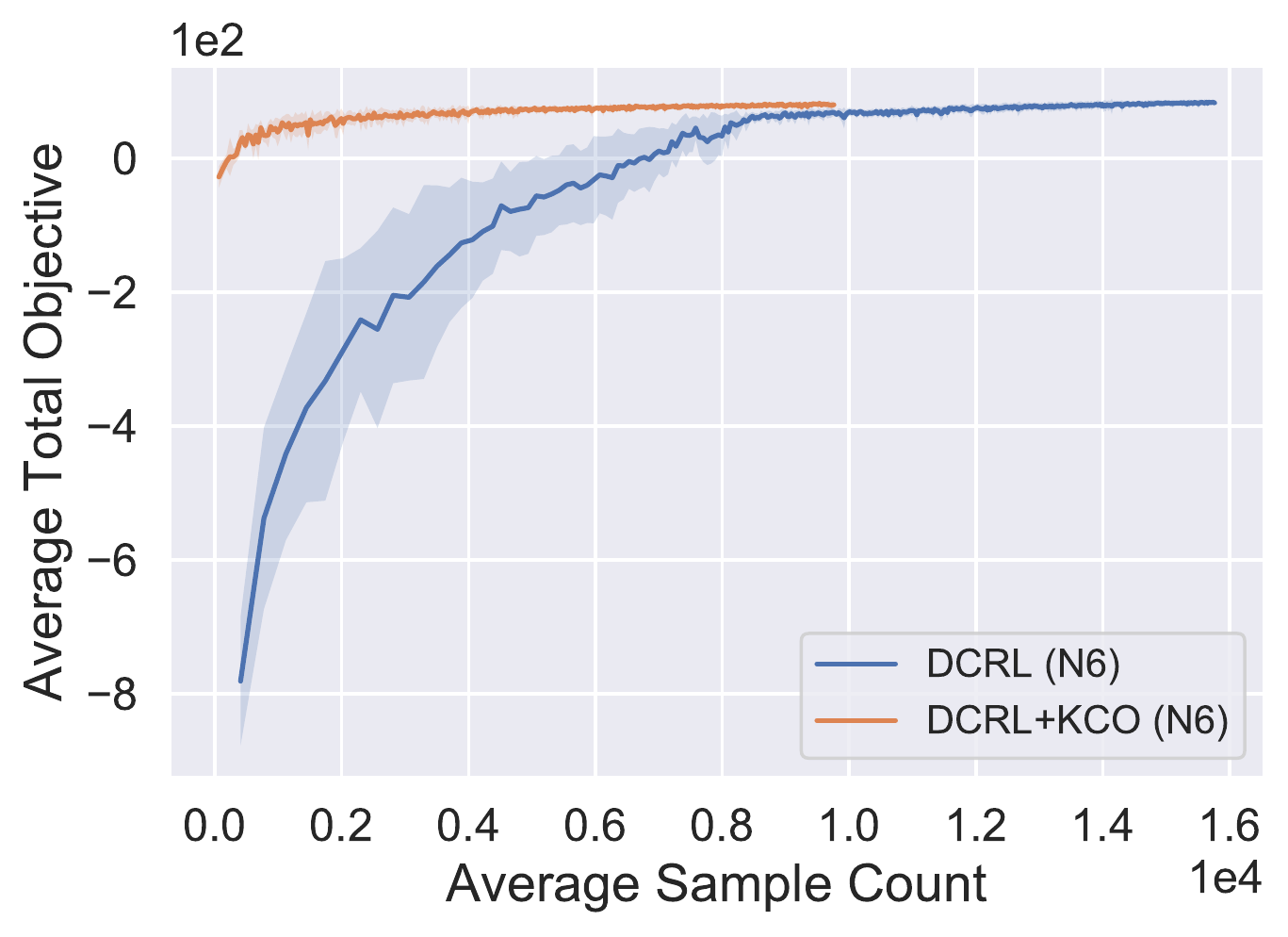}}
	\subfigure[10x10 grid]{\includegraphics[scale=0.4]{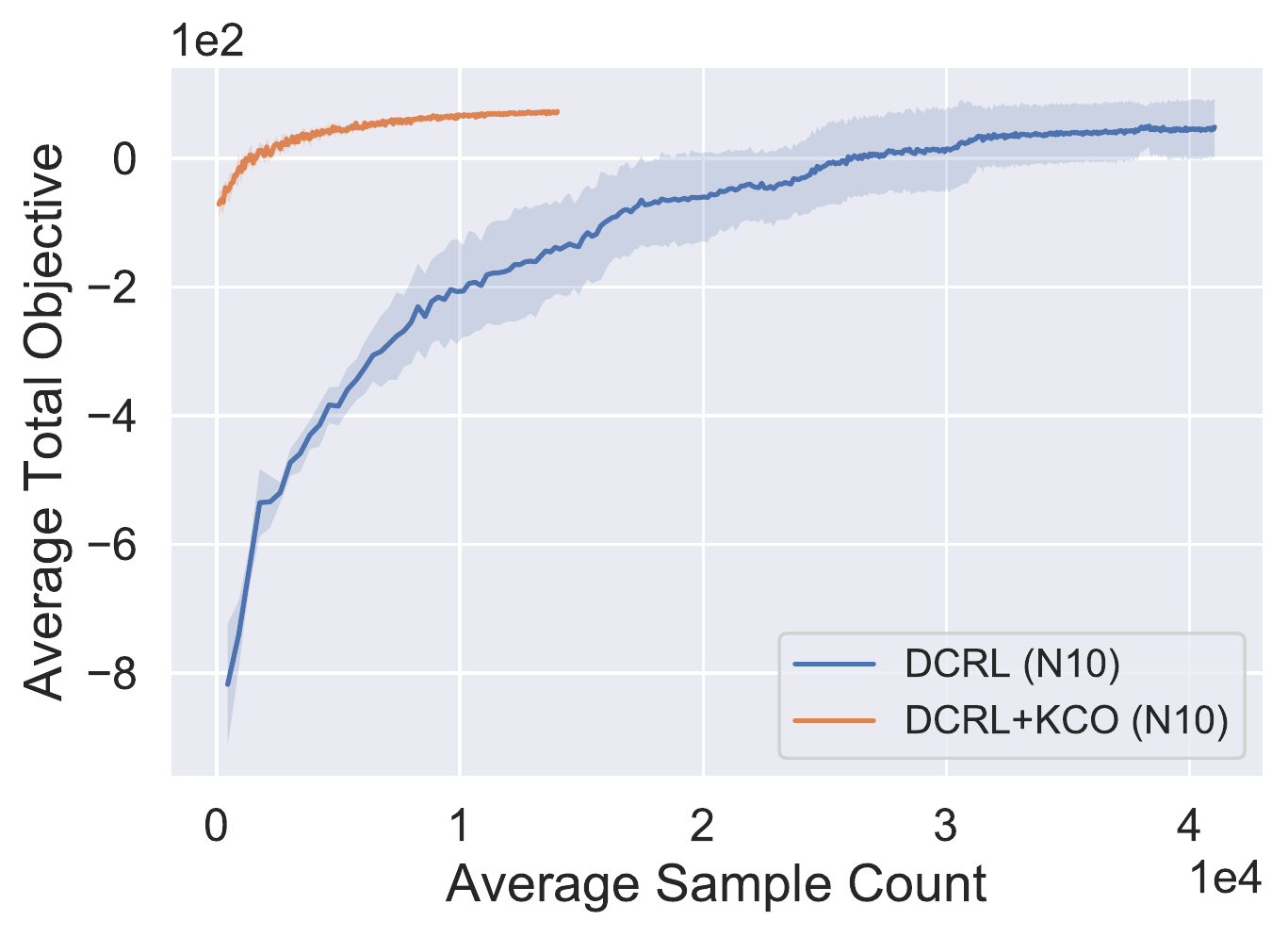}}
	\vskip -10pt
	\caption{\small Sample efficiency comparison between DCRL+KCO and DCRL on open grids (N\# denotes number of agents)}
	\label{fig:DCRL_open1}
	\vskip -10pt
\end{figure}

\begin{figure}[H]
	\centering
	\subfigure[4x4 grid]{\includegraphics[scale=0.4]{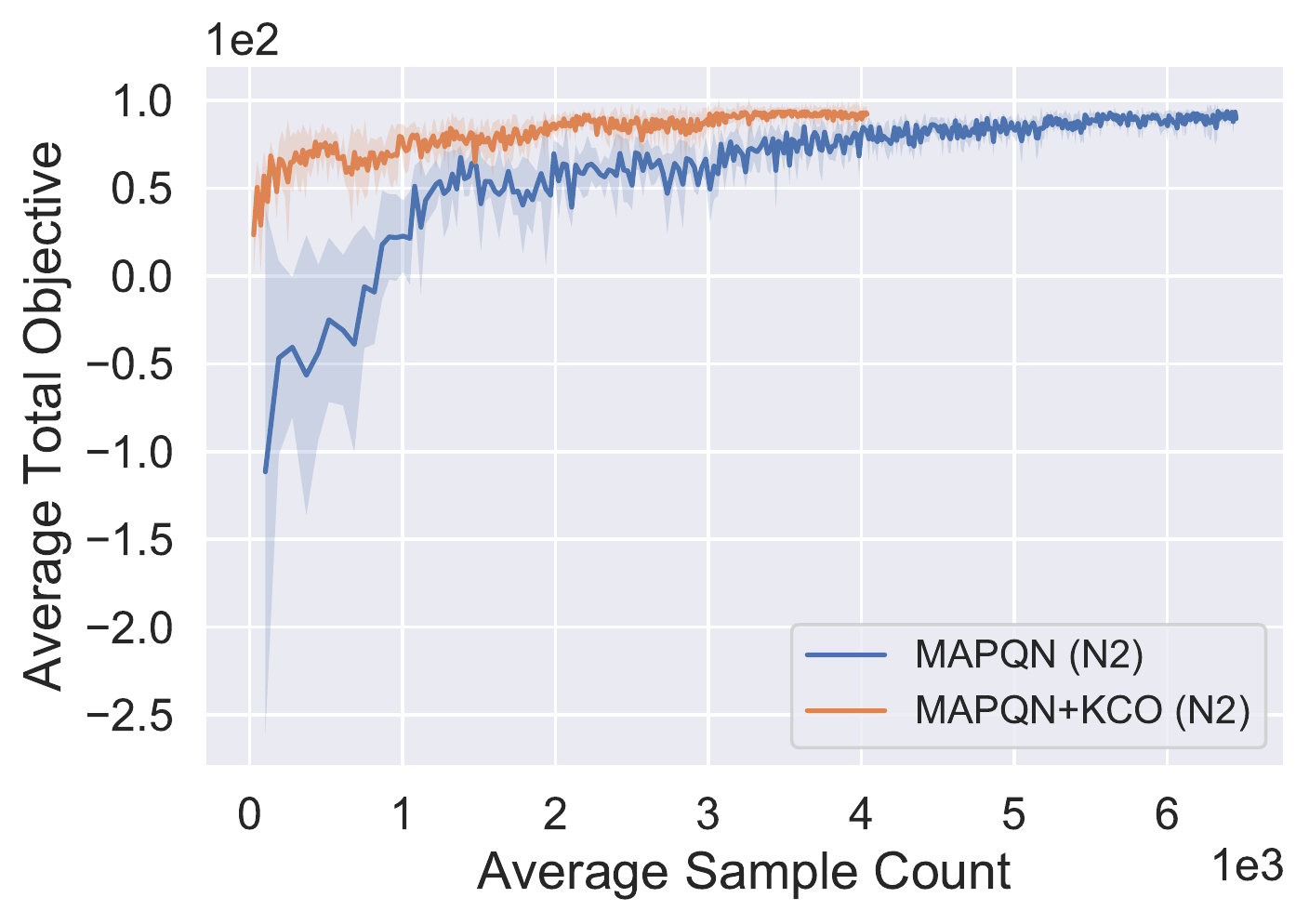}}
	\subfigure[8x8 grid]{\includegraphics[scale=0.4]{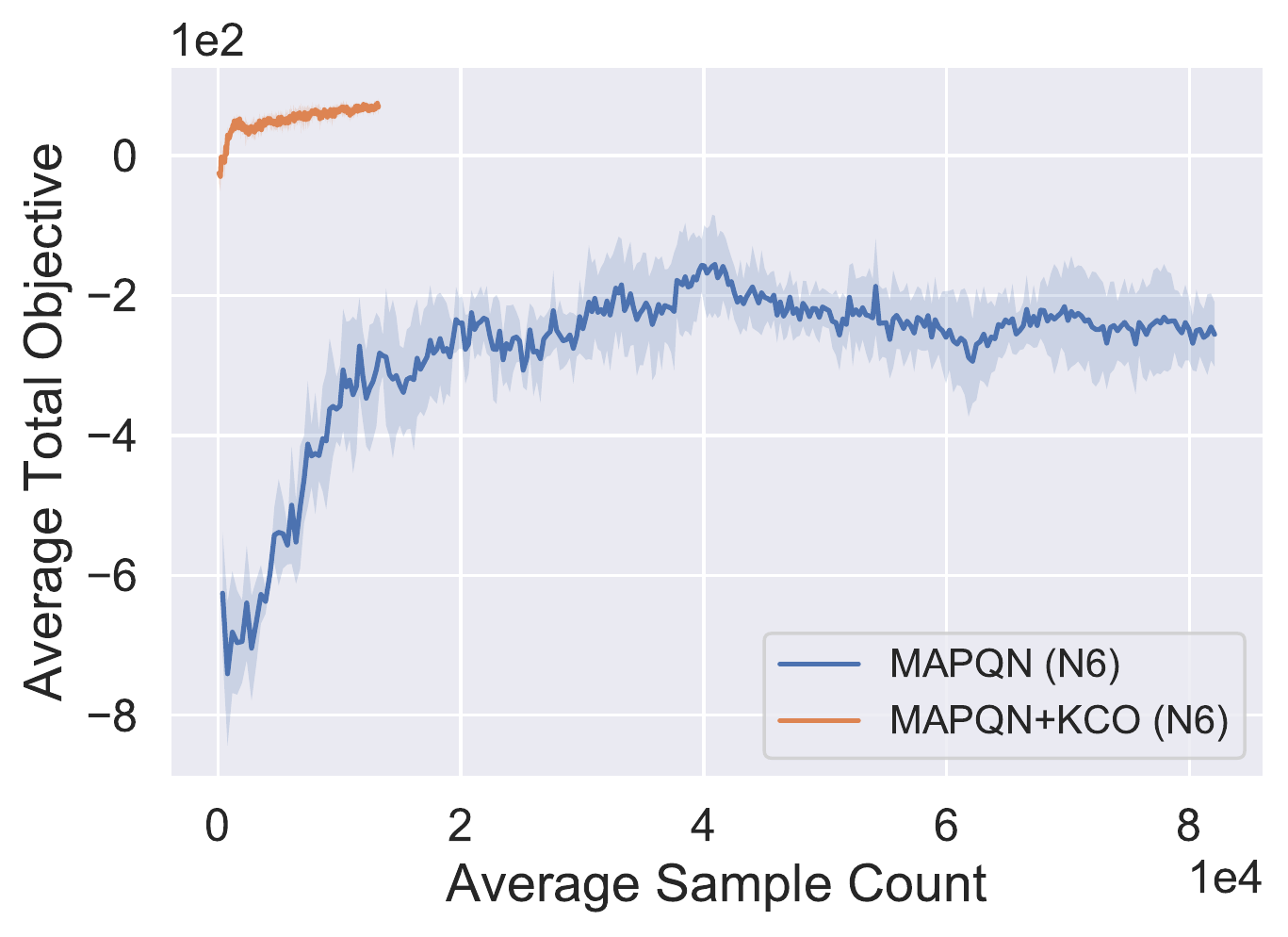}}
	\vskip -10pt
	\caption{\small Sample efficiency comparison between MAPQN+KCO and MAPQN on open grids (higher quality better)}
	\label{fig:MAPQN_open1}
	\vskip -10pt
\end{figure}

\begin{figure}[H]
	\centering
	\subfigure[DCRL+KCO]{\includegraphics[scale=0.4]{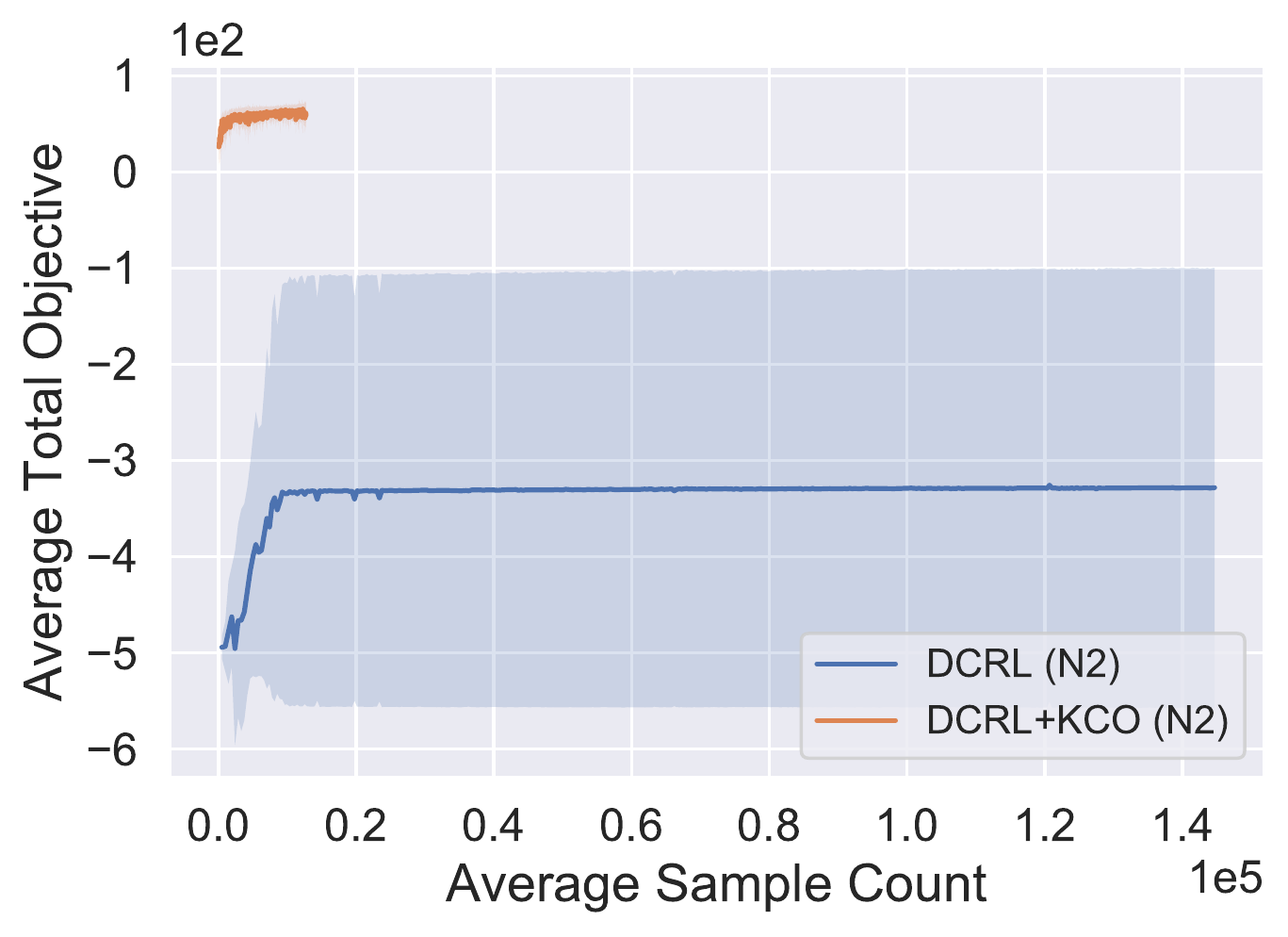}}
	\subfigure[MAPQN+KCO]{\includegraphics[scale=0.4]{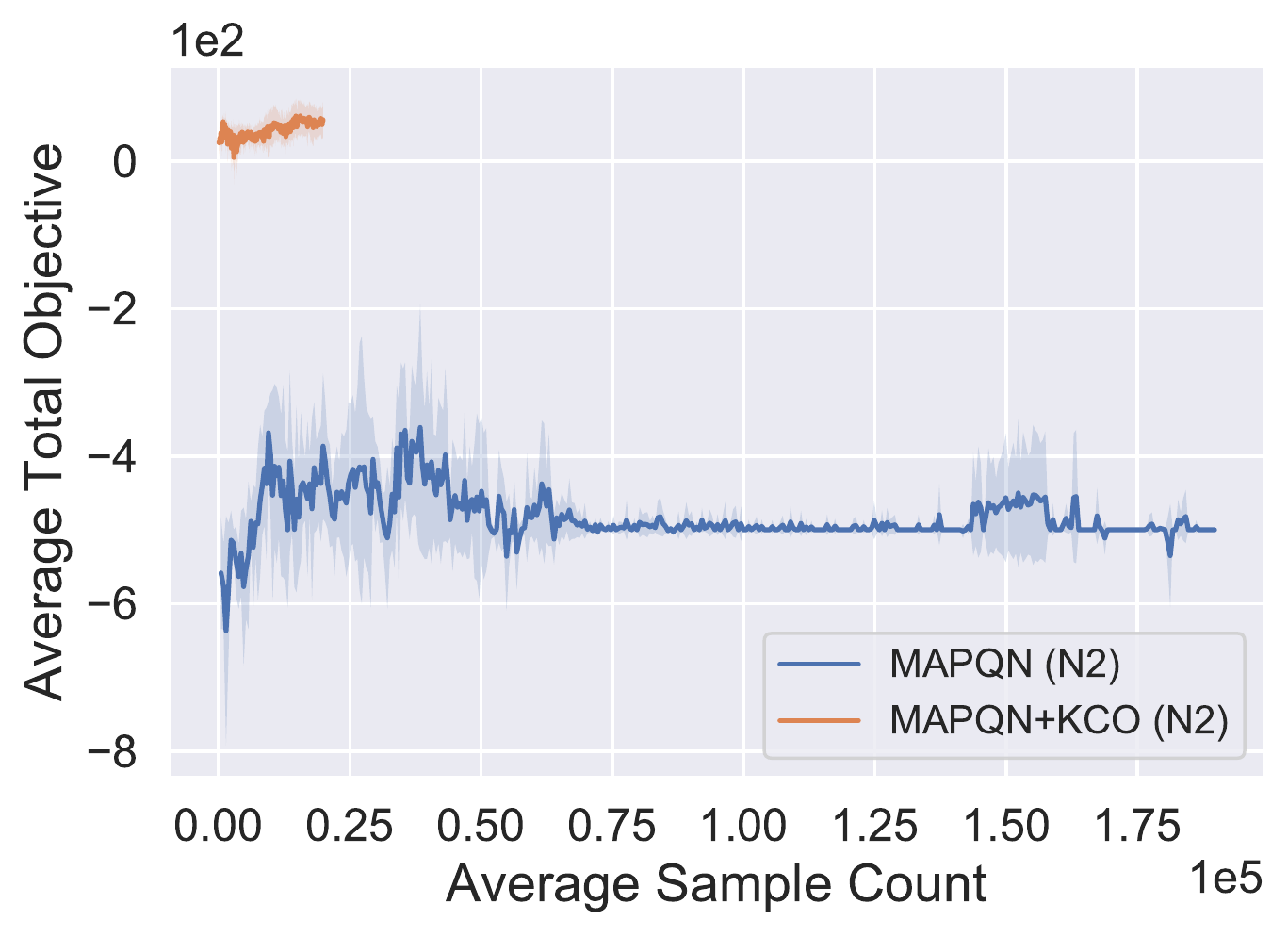}}
	\vskip -10pt
	\caption{\small Sample efficiency results on 10x10 grid with obstacles}
	\label{fig:10x10_ob}
	\vskip -10pt
\end{figure}

\noindent{\textbf{Average Total Objective: }}We show the plots of average total objective vs average sample count for different settings. Figure~\ref{fig:DCRL_open1} shows the results for 4x4 with 2 agents, 8x8 with 6 agents, and 10x10 with 10 agents by DCRL and DCRL+KCO. We clearly observe that DCRL+KCO converges much faster than DCRL especially on the 10x10 grid. Figure~\ref{fig:MAPQN_open1} shows the comparison of MAPQN and MAPQN+KCO for 4x4 with 2 agents and 8x8 with 6 agents. Again, MAPQN+KCO is more sample efficient than MAPQN. The solution quality is better by MAPQN+KCO as well since all the agents are able to reach their respective destinations. Figure~\ref{fig:10x10_ob} shows the results of different approaches on 10x10 grid with obstacles. It clearly shows that DCRL+KCO and MAPQN+KCO are performing much better.

\noindent{\textbf{Stranded Agents: }}Table~\ref{tab:straned_agents} shows the stranded agents on different settings. All agents can reach their destinations on all experimental settings by DCRL+KCO and MAPQN+KCO. DCRL performs reasonably well on open grids in terms of stranded agents. However, 
several agents did not reach the destination on 10x10 grid with obstacles by DCRL. MAPQN performs badly especially on 10x10 grid with obstacles.

\begin{table}[H]
	\centering
	\resizebox{0.8\linewidth}{!}{
	\begin{tabular}{|l| p{3cm}  p{3cm} |p{3cm}  p{3cm}| }
		\hline
	\textbf{Setting}         & \textbf{DCRL} &\textbf{DCRL+KCO} &\textbf{MAPQN}	&\textbf{MAPQN+KCO} \\		
		\hline
		4x4 N2 		& 0    & 0	 	& 0      & 0\\
		4x4 N4		& 0    & 0	 	& 0      & 0\\
		4x4 N6 		& 0    & 0	 	& 0      & 0\\
		8x8 N6		& 0    & 0	    & 3.6    & 0\\
		8x8 N12		& 0    & 0	 	& 9.8    & 0\\
		8x8 N20		& 0    & 0	 	& 12.4   & 0\\
		10x10 N10	& 0.2  & 0	    & - 	 & - \\
		10x10 N20	& 0    & 0		& - 	 & - \\
		10x10 N30	& 0.6  & 0	    & - 	 & - \\
		10x10 with obstacles N2	    & 1.4  & 0		& 2	      & 0\\
		10x10 with obstacles N5	    & 1	   & 0		& 5       & 0\\
		10x10 with obstacles N10	& 2.2  & 0		& 8.8     & 0\\
		\hline
	\end{tabular}}
\caption{\small Average stranded agents comparisons on different settings (N\# denotes number of agents)}
\label{tab:straned_agents}
\end{table}

% \clearpage

% \small
%\bibliographystyle{aaai}
\bibliography{references}
\end{document}